\pdfoutput=1  % force pdfLaTeX on arXiv (source contains PDF and JPEG figures)
\documentclass{article}

\usepackage{arxiv}

\usepackage[utf8]{inputenc}
\usepackage[T1]{fontenc}
\usepackage[hyphens]{url}
\usepackage{hyperref}
\usepackage{graphicx}
\usepackage{natbib}
\usepackage{caption}
\usepackage{amsmath,amssymb}
\usepackage{booktabs}
\usepackage{microtype}
\usepackage{tikz}
\usetikzlibrary{arrows.meta,positioning,fit}

\definecolor{phOne}{HTML}{1F6FB2}
\definecolor{phOneBg}{HTML}{EDF4FB}
\definecolor{phTwo}{HTML}{C05B12}
\definecolor{phTwoBg}{HTML}{FDF3E8}
\definecolor{hand}{HTML}{2F7D6B}
\definecolor{handBg}{HTML}{EAF5F1}
\definecolor{ink}{HTML}{3C4043}

\makeatletter
\renewcommand{\fps@figure}{htbp}
\renewcommand{\fps@table}{htbp}
\makeatother
\title{Neurosymbolic Embodied Agents}
\author{
  Mohammad Albinhassan \\
  Department of Computing \\
  Imperial College London \\
  \texttt{m.albinhassan23@imperial.ac.uk}
  \And
  Yuming Feng \\
  Department of Computer Science \\
  Johns Hopkins University, Whiting School of Engineering \\
  \texttt{yfeng97@jh.edu}
  \AND
  Alessandra Russo \\
  Department of Computing \\
  Imperial College London \\
  \texttt{a.russo@imperial.ac.uk}
  \And
  Pranava Madhyastha \\
  Department of Computer Science \\
  City, University of London \\
  \texttt{pranava.madhyastha@city.ac.uk}
}
\date{}

\renewcommand{\shorttitle}{Neurosymbolic Embodied Agents}
\hypersetup{
  pdftitle={Neurosymbolic Embodied Agents},
  pdfauthor={Mohammad Albinhassan, Yuming Feng, Alessandra Russo, Pranava Madhyastha}
}

\begin{document}
\maketitle

\begin{abstract}

Language and vision-language models generate plausible embodied plans but do not guarantee executability, as their outputs can violate environment dynamics or act on incorrectly grounded entities. We present a neurosymbolic agent that factors long-horizon household tasks into task-directed visual exploration and constrained symbolic planning. In the first phase, a vision-language model and exploration harness acquire goal-relevant predicates and instance bindings from egocentric observations and grounded interactions, producing a symbolic initial state. In the second, a PDDL transition model restricts decoding to tokens that extend applicable actions. Monte Carlo tree search then evaluates executable continuations using a domain-independent planning heuristic. The resulting plans are executable by construction under the transition model, with transfer to the environment conditioned on correct visual grounding. On VirtualHome and ALFWorld, open 4B--27B models exceed 90\% success in both environments, and our smallest agent substantially outperforms a 27B direct visual policy in each. Constraints and search prove complementary rather than interchangeable: in ALFWorld either alone solves under a third of tasks, whereas their combination solves over 95\%. The method also uses several times fewer generated tokens than extended thinking and far fewer model-visible images than direct interaction, and residual failures localize to state acquisition rather than plan generation without any specialized training.
  \end{abstract}

\section{Introduction}

Vision-language models (VLMs) provide a appealing interface for embodied agents. They recognize objects, draw on commonsense knowledge about environments, and generate plausible action sequences from natural-language goals. These capabilities suggest a direct route from perception to control, i.e., show the model what the agent sees, request the next action, execute it, and repeat. In the context of long-horizon tasks, however, plausibility is likely no sufficient. A single action applied to the wrong instance, before its preconditions hold, or under an incorrect belief about the scene can invalidate the remaining trajectory. Such errors accumulate over interaction, and neither model scale nor fluent reasoning guarantees that the final sequence is executable~\citep[inter alia]{valmeekam2022planbench,kambhampati2024llmscantplan,valmeekam2025a,he2025vlmformalize}.

This limitation reflects a difficult coupling of perception and planning.
From an egocentric image, an agent must identify task-relevant objects,
determine their spatial relations and states, preserve these facts across
viewpoints, select applicable actions, and reason about their delayed
effects. A direct policy must solve all of these inside one autoregressive
history, where they are free to compound: a fluent plan may reference an
unobserved object, and a correct observation may still be followed by an
inapplicable action. Execution feedback can expose such an error after it
occurs, but does not prevent the model from repeating an invalid choice or
drifting from the state its own actions induced. Existing embodied systems
mitigate parts of this through affordance models, replanning, or
task-and-motion
planning~\citep{ahn2022saycan,huang2023inner,yang2024vlmtamp}. Reliable
high-level task planning nonetheless still requires a mechanism that binds
visually acquired state to formal action semantics during generation
itself.

In this paper, we present a neurosymbolic embodied agent for discrete, long-horizon
household tasks that supplies this mechanism by factoring the problem into
two connected phases (see Figure~\ref{fig:method} for a quick overview). \emph{Phase~I:
Exploration} couples a VLM to an exploration harness that acquires only
the objects, relations, and instance bindings the goal requires. Model
priors guide where to inspect, while multiple viewpoints and grounded
interactions determine which relational claims enter the symbolic state.
\emph{Phase~II: Exploitation} plans from this grounded initial state under
a PDDL transition model. At every decoding step, the model distribution is
restricted to tokens that can extend an applicable action, and Monte Carlo
tree search (MCTS)evaluates executable continuations with a
domain-independent planning heuristic, combining the language model's
procedural prior with explicit state transitions.

The decomposition separates validity from goal reachability.
State-dependent constrained decoding prevents the search from generating
inapplicable actions, while MCTS distinguishes among the remaining
executable plans by their long-horizon consequences. Applicability thereby
becomes an invariant maintained throughout generation and search rather
than a behavior the model must re-infer at each turn, and every returned
plan is executable under the transition model by construction. The
guarantee transfers to the simulator when the symbolic model is sound and
the visually acquired instance binding is correct. This qualification is
important: formal constraints cannot recover an object that Phase~I fails
to perceive. They instead localize the residual uncertainty to state
acquisition, rather than allowing perceptual, grounding, and planning
errors to compound throughout an unconstrained trajectory. Our failure
attribution confirms that this is where the error in fact concentrates.

We evaluate on VirtualHome and ALFWorld, two visually grounded household
benchmarks with different action semantics and grounding demands. Across
Qwen3.5 models from 4B to 27B parameters, the agent obtains
$94.5$--$99.5\%$ success in VirtualHome and $90.3$--$97.8\%$ in ALFWorld.
The 4B agent exceeds the 27B direct visual policy by $32.0$ percentage
points in VirtualHome and $63.4$ points in ALFWorld, indicating that
explicit structure can substitute for considerable model scale. It is also
efficient: with costs paired by model size and averaged across benchmarks,
the method uses up to $4.0\times$ fewer generated tokens than extended
thinking, $1.5\times$ fewer than unconstrained MCTS, and $5.6\times$ fewer
model-visible images than direct interaction.

Our primary contributions are: a) a two-phase architecture that separates task-directed visual state acquisition from goal-directed symbolic planning, through a grounded initial-state interface; b) a state-dependent decoding procedure that enforces PDDL action applicability at token level, coupled with language-guided MCTS over executable plan prefixes, so that plans are executable by construction; and c) an empirical end-to-end evaluation across two embodied household benchmarks,
including direct visual and embodied-model baselines, component ablations, inference and observation costs, and causal failure attribution.

\section{Related Work}

Our work connects symbolic task planning, language-model planning, constrained decoding, and embodied perception to generate executable plans from visual observations.

\paragraph{Symbolic Planning and LLM Planners}
Symbolic task planning searches over explicit world states. PDDL is the dominant modeling language~\citep{mcdermott1998pddl,fox2003pddl21} where grounding predicates and operators yields a task whose actions are executable only when their preconditions hold, after which their effects update the state. This enables exact feasibility checks, although the combinatorial grounded action space motivates heuristic planners such as Fast Downward~\citep{helmert2006fastdownward}. We use the same applicability semantics online, to constrain LLM decoding and MCTS. LLMs supply useful procedural priors but do not reliably enforce these semantics and it has been shown that fluency does not imply correct reasoning over preconditions, effects, and state change~\citep[inter alia]{valmeekam2022planbench}, and even frontier models require external validation and exhibit hallucinated actions or unstable plans when problems are expressed in language~\citep{kambhampati2024llmscantplan,correa2025frontier,armony2025symbolic,katz2026planning}.

\paragraph{LLMs with PDDL and Symbolic Planners}
LLM+P and NL2Plan translate language into PDDL and invoke a planner, incrementally constructing domains and problems from minimal descriptions~\citep{liu2023llmp,gestrin2024nl2plan}, but generated domains can be syntactically plausible yet semantically incorrect~\citep{oswald2024domaingenerators,smirnov2024consistent,zhang2024proc2pddl}. Later systems iteratively refine PDDL with agents and verifiers, expose symbolic transitions as tools, or induce action models from demonstrations~\citep{lamalfa2025endtoend,gobel2026stepwise,huang2025pddllm}. These approaches largely treat PDDL as a generated representation or downstream solver input. We instead expose grounded operator applicability \emph{during} inference, rejecting invalid continuations and restricting tree search to executable prefixes.

\paragraph{Constrained Decoding for Structured Generation}
Constrained decoding masks next tokens that violate a formal specification, supporting structured generation without fine-tuning and with efficient, reliable handling of subword tokenization, schema evaluation, and dynamic agent outputs~\citep{geng2023grammar,beurerkellner2024domino,ugare2024syncode,dong2024xgrammar,geng2025jsonschemabench,park2025flexible,li2026xgrammar2}. Most of this work addresses syntax or text-only semantic control. Closest to ours, \texttt{SEM-CTRL}  couples controlled decoding with search under an external transition model to enforce semantic validity, and later work learns such constraints automatically under an oracle~\citep{albinhassan2026semctrl,albinhassan-etal-2025-learning}. We extend state-dependent constraints to visually grounded planning, where the state the constraints act on is not given but must first be acquired through perception and interaction.

\paragraph{Perception and Planning}
Separating representation construction from goal-directed control is a long-standing design pattern in robotics: SLAM estimates a reusable environment model for subsequent navigation~\citep{cadena2016slam}, while modular agents combine learned mapping, semantic priors over likely goal locations, and online scene graphs with analytic planning and exploration-objective selection~\citep{chaplot2020slam,chaplot2020semexp,yokoyama2024vlfm,yin2025unigoal}. The shared division of labor is that semantic knowledge guides where to acquire information, whereas grounded observations determine the represented state. Language-conditioned systems apply the same split, pairing LLM proposals with affordances, tools, or execution feedback~\citep{ahn2022saycan,yao2022react,huang2023inner}, and VLM-TAMP uses a VLM to propose horizon-reducing subgoals while a task-and-motion planner enforces geometric and kinematic feasibility~\citep{yang2024vlmtamp}. A systematic study of multimodal planning reaches a complementary conclusion where we see that VLMs perform substantially better as PDDL formalizers than as end-to-end long-horizon planners, with incomplete visual extraction of object relations the principal bottleneck~\citep{he2025vlmformalize}. This directly motivates our decomposition of visual state acquisition from constrained symbolic planning.

\section{Method: Neurosymbolic Embodied Agents}
\label{sec:method}

We consider long-horizon embodied task planning at the level of discrete, semantically meaningful actions, rather than continuous control or motion generation. These tasks couple two distinct challenges: a) an agent must first determine what is present and where it is, then b) synthesize a sequence of actions that achieves the goal. A model must resolve occlusion, containment, and entity grounding before reasoning over action preconditions and long-horizon effects. Solving both in one autoregressive loop entangles perceptual and planning errors where a fluent plan may reference an unobserved object, while a correct observation may still be followed by an inapplicable action.
We factor this tightly coupled problem into two explicitly connected subproblems. During \emph{Phase~I: Exploration}, a vision--language model (VLM) actively acquires the scene facts needed by the task. During \emph{Phase~II: Exploitation}, a language model plans from the resulting symbolic representation while a formal transition model constrains every action. The interface assigns state acquisition to Phase~I and valid, goal-directed deliberation to Phase~II.

\paragraph{Problem setting.}
We model each task as a partially observable Markov decision process (POMDP) $\mathcal{M}=\langle\mathcal{S},\mathcal{A},T,R,\Omega,O,\gamma\rangle$ \citep{kaelbling1998planning}.  Here, $\mathcal{S}$ and $\mathcal{A}$ are the state and action spaces; $T(s'\mid s,a)$ is the transition model; $\Omega$ is the space of egocentric visual observations; and $O(I\mid s)$ is the observation model.  The sparse reward $R$ is positive only when the resulting state satisfies goal formula $G$.  We consider a finite, undiscounted horizon ($\gamma=1$).

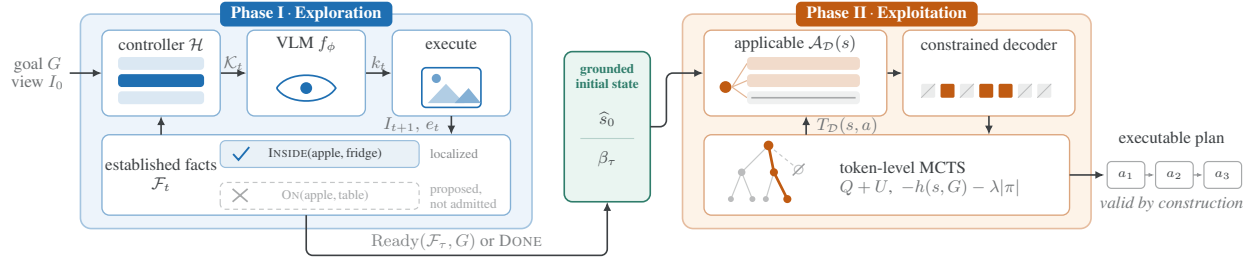
\begin{figure*}[t]
  \centering
  \resizebox{\textwidth}{!}{
\begin{tikzpicture}[
   x=1mm,y=1mm,font=\small,text=ink,
   flow/.style={-{Latex[length=1.7mm,width=1.5mm]},draw=ink,line width=0.3mm,rounded corners=1.2mm},
   lab/.style={font=\footnotesize,text=ink!75,inner sep=0.5mm},
   ttl/.style={font=\small,text=ink,inner sep=0pt},
   b1/.style={draw=phOne!55,fill=white,line width=0.25mm,rounded corners=1.4mm},
   b2/.style={draw=phTwo!55,fill=white,line width=0.25mm,rounded corners=1.4mm},
   chip/.style={rounded corners=1mm,inner xsep=2mm,inner ysep=0.9mm,font=\small\bfseries,text=white}]
 \draw[rounded corners=2mm,fill=phOneBg,draw=phOne!40,line width=0.3mm] (12,12) rectangle (96,52);
 \draw[rounded corners=2mm,fill=phTwoBg,draw=phTwo!40,line width=0.3mm] (124,12) rectangle (200,52);
 \node[chip,fill=phOne] at (54,52) {Phase I\,$\cdot$\,Exploration};
 \node[chip,fill=phTwo] at (162,52) {Phase II\,$\cdot$\,Exploitation};
 \node[align=center,font=\footnotesize,text=ink!75] at (4,41) {goal $G$\\ view $I_0$};
 \node[b1,minimum width=22mm,minimum height=16mm] (proto) at (27,41) {};
 \node[ttl] at (27,46.4) {controller $\mathcal{H}$};
 \foreach \y in {41.6,35.2}{\fill[phOne!16,rounded corners=0.7mm] (19,\y) rectangle ++(16,2.4);}
 \fill[phOne,rounded corners=0.7mm] (19,38.4) rectangle ++(16,2.4);
 \node[b1,minimum width=22mm,minimum height=16mm] (vlm) at (54,41) {};
 \node[ttl] at (54,46.4) {VLM $f_\phi$};
 \draw[phOne,line width=0.32mm] (48.5,38.2) .. controls (50.5,41.4) and (57.5,41.4) .. (59.5,38.2)
        .. controls (57.5,35) and (50.5,35) .. (48.5,38.2);
 \fill[phOne] (54,38.2) circle (1.1mm);
 \node[b1,minimum width=22mm,minimum height=16mm] (act) at (81,41) {};
 \node[ttl] at (81,46.4) {execute};
 \draw[phOne,line width=0.3mm,rounded corners=0.8mm] (75.5,34.6) rectangle (86.5,42);
 \fill[phOne!65] (78.2,40) circle (0.85mm);
 \fill[phOne!35] (76.6,35.4) -- (80,39) -- (83.4,35.4) -- cycle;
 \fill[phOne!55] (81.6,35.4) -- (84.2,38.2) -- (86.5,35.4) -- cycle;
 \node[b1,minimum width=76mm,minimum height=15mm] (card) at (54,22) {};
 \node[align=center,font=\footnotesize,text=ink] at (27,22) {established facts\\ $\mathcal{F}_t$};
 \draw[rounded corners=1mm,fill=phOne!10,draw=phOne!45,line width=0.22mm] (38,23.4) rectangle (75,28.4);
 \node[font=\scriptsize] at (57.5,25.9) {\textsc{Inside}(apple,\,fridge)};
 \draw[phOne,line width=0.35mm] (40.4,25.9) -- (41.4,24.8) -- (43.4,27.2);
 \draw[rounded corners=1mm,fill=white,draw=ink!30,line width=0.22mm,dash pattern=on 0.9mm off 0.7mm]
       (38,15.6) rectangle (75,20.6);
 \node[font=\scriptsize,text=ink!45] at (57.5,18.1) {\textsc{On}(apple,\,table)};
 \draw[ink!45,line width=0.32mm] (40.3,17) -- (43.1,19.4) (43.1,17) -- (40.3,19.4);
 \node[anchor=west,inner sep=0pt,font=\scriptsize,text=ink!60] at (77,25.9) {localized};
 \node[anchor=west,inner sep=0pt,align=left,font=\scriptsize,text=ink!45] at (77,18.1) {proposed,\\ not admitted};
 \draw[rounded corners=1.5mm,fill=handBg,draw=hand,line width=0.3mm] (102,17) rectangle (118,45);
 \node[align=center,font=\scriptsize\bfseries,text=hand] at (110,40.5) {grounded\\initial state};
 \node[font=\small] at (110,32.5) {$\widehat{s}_0$};
 \node[font=\small] at (110,25) {$\beta_\tau$};
 \draw[hand!35,line width=0.2mm] (105,28.8) -- (115,28.8);
 \node[b2,minimum width=34mm,minimum height=16mm] (pddl) at (145,41) {};
 \node[ttl] at (145,46.4) {applicable $\mathcal{A}_{\mathcal{D}}(s)$};
 \fill[phTwo] (132,38.4) circle (1.1mm);
 \foreach \y in {41.6,38.4,35.2}{\draw[phTwo!45,line width=0.22mm] (133.2,38.4) -- (136,{\y+1.2});}
 \fill[phTwo!22,rounded corners=0.7mm] (136,41.6) rectangle ++(21,2.4);
 \fill[phTwo!22,rounded corners=0.7mm] (136,38.4) rectangle ++(21,2.4);
 \fill[ink!8,rounded corners=0.7mm] (136,35.2) rectangle ++(21,2.4);
 \draw[ink!45,line width=0.3mm] (136.8,36.4) -- (156.2,36.4);
 \node[b2,minimum width=32mm,minimum height=16mm] (dec) at (181,41) {};
 \node[ttl] at (181,46.4) {constrained decoder};
 \foreach \i in {1,3,4}{\fill[phTwo,rounded corners=0.3mm] ({168.4+\i*3.6},36.4) rectangle ++(2.6,2.6);}
 \foreach \i in {0,2,5,6}{\fill[ink!10,rounded corners=0.3mm] ({168.4+\i*3.6},36.4) rectangle ++(2.6,2.6);
   \draw[ink!35,line width=0.16mm] ({168.4+\i*3.6},36.4) -- ++(2.6,2.6);}
 \node[b2,minimum width=68mm,minimum height=15mm] (mcts) at (162,22) {};
 \draw[ink!28,line width=0.22mm] (140,27.5) -- (135,22.5) (135,22.5) -- (133,17.5)
       (135,22.5) -- (137,17.5) (141,22.5) -- (139.5,17.5);
 \draw[ink!28,line width=0.22mm,dash pattern=on 0.7mm off 0.6mm] (140,27.5) -- (145.1,23.4);
 \draw[ink!30,line width=0.22mm,fill=white] (145.6,23) circle (0.8mm);
 \draw[ink!50,line width=0.28mm] (144.5,21.9) -- (146.7,24.1);
 \draw[phTwo,line width=0.5mm] (140,27.5) -- (141,22.5) -- (143.5,17.5);
 \fill[ink!35] (135,22.5) circle (0.7mm) (133,17.5) circle (0.55mm)
               (137,17.5) circle (0.55mm) (139.5,17.5) circle (0.55mm);
 \fill[phTwo] (140,27.5) circle (0.9mm) (141,22.5) circle (0.75mm) (143.5,17.5) circle (0.75mm);
 \node[anchor=west,align=left] at (152,23.4) {token-level MCTS};
 \node[anchor=west,align=left,font=\footnotesize,text=ink!75] at (152,19.6) {$Q+U$,\ \ $-h(s,G)-\lambda|\pi|$};
 \node[align=center,font=\footnotesize,text=ink] at (215,28.6) {executable plan};
 \foreach \i/\t in {0/1,1/2,2/3}{
   \draw[rounded corners=0.8mm,fill=white,draw=ink!45,line width=0.22mm]
        ({203+\i*9},19.5) rectangle ++(7,5);
   \node[font=\scriptsize] at ({206.5+\i*9},22) {$a_{\t}$};}
 \draw[-{Latex[length=1.1mm]},ink!55,line width=0.22mm] (210,22) -- (211.8,22);
 \draw[-{Latex[length=1.1mm]},ink!55,line width=0.22mm] (219,22) -- (220.8,22);
 \node[align=center,font=\footnotesize\itshape,text=ink!70] at (215,16.6) {valid by construction};
 \draw[flow] (10,41) -- (16,41);
 \draw[flow] (proto.east) -- node[lab,above]{$\mathcal{K}_t$} (vlm.west);
 \draw[flow] (vlm.east) -- node[lab,above]{$k_t$} (act.west);
 \draw[flow] (81,33) -- (81,29.5);
 \node[lab,anchor=east] at (79.5,31.3) {$I_{t+1},\,e_t$};
 \draw[flow] (27,29.5) -- (27,33);
 \draw[flow] (54,14.5) -- (54,7) -- (110,7) -- (110,17);
 \node[lab,anchor=south,text=ink!65] at (82,7.4) {$\operatorname{Ready}(\mathcal{F}_\tau,G)$ or \textsc{Done}};
 \draw[flow] (118,31) -- (121,31) |- (pddl.west);
 \draw[flow] (pddl.east) -- (dec.west);
 \draw[flow] (181,33) -- (181,29.5);
 \draw[flow] (147,29.5) -- (147,33);
 \node[lab,anchor=west] at (148.4,31.2) {$T_{\mathcal{D}}(s,a)$};
 \draw[flow] (196,22) -- (202.5,22);
\end{tikzpicture}}
  \caption{\textbf{Two phases joined by a narrow symbolic interface.} A literal enters $\mathcal{F}_t$ only under visual or interaction evidence, so priors decide \emph{where to look}, not \emph{what is present} (greyed chip: unsupported). Only $\widehat{s}_0$ and $\beta_\tau$ cross into Phase~II, which excludes at three scales: inapplicable actions (struck pill), illegal tokens (slashed cells), unexpanded branches (struck node).}  \label{fig:method}
\end{figure*}

At time $t$, the agent receives only an egocentric RGB observation $I_t\in\Omega$, while the complete environment state $s_t\in\mathcal{S}$ remains latent, we note that objects may be outside the field of view, occluded, or enclosed. The output is a plan $\pi=(a_1,\ldots,a_n)$ whose actions are sequentially applicable and whose final state satisfies the goal, written $s_n\models G$. We do not solve the POMDP in belief space and emphasize that Phase~I acts to reduce uncertainty over the task-relevant state and commits to a point estimate, after which Phase~II solves a deterministic classical planning problem and returns an open-loop plan. This reduction is exact when the dynamics reachable from that estimate are deterministic and the estimate is correct on the literals the plan depends on conditions we make precise in the validity guarantee below. Generative models provide useful plan priors but do not ensure accurate state estimation or applicability. We present a higher-level overview in Figure~\ref{fig:method}.

\paragraph{Goal specifications.} The goal formula $G$ is obtained by a deterministic, rule-based parse of each benchmark's own symbolic goal annotation, applied uniformly to every episode (we explain more in Experimental Setup). We note that this is not hand-specified and has only class-specific information (hence no potential for leakage of information).

\subsection{Phase I (Exploration): Task-Directed Visual Grounding}
\label{sec:exploration}
To avoid exhaustive scene reconstruction, a VLM, denoted by $f_\phi$ with parameters $\phi$, is coupled to a deterministic exploration controller $\mathcal{H}$ (harness). The VLM selects inspection targets from egocentric frames, while the harness tracks inspection history, grounds skills, and enforces environment interaction constraints. This is shown in the left half of Figure~\ref{fig:method}. We now formalize the exploration phase below.

Let $\mathcal{L}$ be the set of possible grounded literals, such as $\textsc{Inside}(\textit{apple},\textit{fridge})$. After $t$ exploration steps, the established facts form $\mathcal{F}_t\subseteq\mathcal{L}$; $\mathcal{R}_t$ records inspected locations, attempted skills, and outcomes. The controller function $\mathcal{H}$ constructs candidate skills
\begin{equation}
  \begin{aligned}
  \mathcal{K}_t&=\mathcal{H}(G,\mathcal{F}_t,\mathcal{R}_t)\\
  &\subseteq \{\textsc{Goto}(x),\textsc{Open}(x),
              \textsc{Close}(x),\textsc{Look},\textsc{Done}\}.
  \end{aligned}
  \label{eq:protocol}
\end{equation}
Here $x$ denotes a navigable object, surface, or receptacle. The skill family is fixed, while $\mathcal{K}_t$ is a state-dependent subset because targets that have already been inspected or are not yet grounded may be omitted. \textsc{Look} is an active inspection macro: it acquires additional viewpoints and, when possible, invokes a grounded interaction to test an object--receptacle hypothesis. Such interactions, including temporary pickup and restoration, are internal grounding operations rather than VLM-selected task actions. \textsc{Done} terminates exploration. The harness may require another inspection or container-opening action when evidence is insufficient, but it does not predict unseen contents.

Given the goal, current image, established facts, interaction record, and candidate skills, it proposes
\begin{equation}
  (\mathcal{P}_t,k_t)
  =f_\phi(G,I_t,\mathcal{F}_t,\mathcal{R}_t,\mathcal{K}_t),
  \qquad k_t\in\mathcal{K}_t.
  \label{eq:vlm-explore}
\end{equation}
Here $\mathcal{P}_t$ is the set of facts proposed from the current image, such as $\textsc{Visible}(o)$, $\textsc{On}(o,r)$, or $\textsc{Inside}(o,r)$, and $k_t$ is the selected exploration skill. The symbols $o$ and $r$ denote an object and a supporting surface or receptacle, respectively.

The environment grounds and executes $k_t$, returning image $I_{t+1}$ and event $e_t$. The event records success or failure and, on failure, a concise grounding error. The evidence update is
\begin{equation}
  (\mathcal{F}_{t+1},\mathcal{R}_{t+1},\beta_{t+1})
  =\mathcal{U}(\mathcal{F}_t,\mathcal{R}_t,\beta_t,
  \mathcal{P}_t,k_t,I_{t+1},e_t),
  \label{eq:fact-update}
\end{equation}
where $\beta_t$ maps symbolic names to environment instances and $\mathcal{U}$ is a deterministic evidence-update function. It adds action effects only after successful execution and commits object--receptacle bindings only when supported by localization from multiple viewpoints or by a successful grounded interaction. Unsupported or conflicting proposals may guide another inspection but do not establish a relational fact. This is not a general hallucination detector, so visual recognition errors can still cause Phase~I failure.

Let $\tau$ denote the final exploration step. Exploration terminates when $\operatorname{Ready}(\mathcal{F}_\tau,G)$ is true, when the model selects \textsc{Done}, or when the interaction budget is exhausted. The Boolean function $\operatorname{Ready}$ tests whether the objects and relations needed to instantiate the planning problem have been grounded. Phase~I returns the grounded initial state
\begin{equation}
  \widehat{s}_0=\mathcal{F}_\tau\cup \mathcal{F}_{\mathrm{str}},
  \qquad \beta_\tau,
  \label{eq:grounded-initial-state}
\end{equation}
where $\mathcal{F}_{\mathrm{str}}$ contains fixed ontology and domain facts shared across tasks, and $\beta_\tau$ is the final symbol-to-instance binding. Neither quantity contains privileged scene state.

\subsection{Phase II (Exploitation): Constrained Symbolic Planning}
\label{sec:exploitation}

As illustrated in the right half of Figure~\ref{fig:method}, Phase~II combines grounded state $\widehat{s}_0$ with a PDDL transition model $\mathcal{D}$, formulating classical planning task $\Pi$. In simple terms, the language model supplies a procedural prior over plans, while PDDL enforces action applicability and tree search evaluates long-horizon consequences. This retains the model's semantic preference without requiring it to infer the transition dynamics from the prompt.

Formally, the problem $\Pi=\langle\mathcal{D},\widehat{s}_0,G\rangle$ comprises domain $\mathcal{D}$, grounded initial state $\widehat{s}_0$, and goal $G$. A grounded operator $a$ has preconditions $\operatorname{pre}(a)$ and effects $\operatorname{add}(a)$ and $\operatorname{del}(a)$. For state $s$, the applicable actions and successor are
\begin{align}
  \mathcal{A}_{\mathcal D}(s)
    &=\{a\mid \operatorname{pre}(a)\subseteq s\}, \label{eq:applicable}\\
  T_{\mathcal D}(s,a)
    &=\bigl(s\setminus\operatorname{del}(a)\bigr)
      \cup\operatorname{add}(a),\quad
      a\in\mathcal{A}_{\mathcal D}(s). \label{eq:transition}
\end{align}
Thus $\mathcal{A}_{\mathcal D}(s)$ contains actions whose preconditions hold, and $T_{\mathcal D}$ computes their successors.  Applicability is available during generation, not only after sampling a plan.

\paragraph{State-dependent constrained decoding.}
Constraints enforce PDDL applicability at token decoding boundaries.
Let $\mathcal{V}$ be the token vocabulary and $\operatorname{tok}(a)$ the tokens spelling action $a$.  At state $s$, let $\mathcal{L}_{\mathcal D}(s)= \{\operatorname{tok}(a)\mid a\in\mathcal{A}_{\mathcal D}(s)\}$ be the applicable action strings, and let $\operatorname{Prefix}(\cdot)$ return all their prefixes.  For partial action $u$, the allowed next tokens are
\begin{equation}
  \mathcal{C}_{\mathcal D}(s,u)=
  \{v\in\mathcal{V}\mid
  u\mathbin{\cdot}v\in\operatorname{Prefix}(\mathcal{L}_{\mathcal D}(s))\},
  \label{eq:constraint}
\end{equation}
where $u\mathbin{\cdot}v$ appends token $v$ to prefix $u$. Let $c_\Pi$ denote the fixed model context for planning problem $\Pi$, including the task goal, grounded initial state, and action interface. The language model with parameters $\theta$ assigns next-token probability $p_\theta(v\mid u,c_\Pi)$. Its state-constrained distribution is
\begin{equation}
  p_{\theta,\mathcal C}(v\mid s,u,c_\Pi)
  \propto p_\theta(v\mid u,c_\Pi)\,
  \mathbb{I}[v\in\mathcal{C}_{\mathcal D}(s,u)],
  \label{eq:constrained-distribution}
\end{equation}
where $\mathbb{I}[\cdot]$ is one when its condition holds and zero otherwise. Thus invalid tokens receive no probability; the remaining model mass is renormalized implicitly.  At each action boundary, $T_{\mathcal D}$ advances the state and the constraint is recomputed.  Unlike a surface-form grammar, the distribution is state dependent: the same action can be permitted in one state and excluded in another.

\paragraph{Validity guarantee.}
Every complete plan generated under distribution $p_{\theta,\mathcal C}$ is applicable under transition model $\mathcal{D}$ by construction. By induction, if a prefix reaches state $s_i$, Equation~\ref{eq:constraint} restricts the subsequent action to $\mathcal{A}_{\mathcal D}(s_i)$, and Equation~\ref{eq:transition} advances the state to $s_{i+1}$. This validity guarantee transfers to the environment provided domain model $\mathcal{D}$ is sound, and Phase~I binding $\beta_\tau$ is accurate.

\paragraph{Constrained language-guided search.}
An applicable sequence may still fail to reach $G$ as validity and goal correctness are distinct \citep[see][for a more detailed discussion]{albinhassan2026semctrl}.  We optimize the latter with token-level MCTS.  A node $n=(s,u)$ contains state $s$ and partial action $u$.  Selection is restricted to $\mathcal{C}_{\mathcal D}(s,u)$ and combines the backed-up search value with the language-model prior:
\begin{equation}
 \begin{aligned}
 v^\star
 &=\arg\max_{v\in\mathcal{C}_{\mathcal D}(s,u)}
   [Q(n,v)+U(n,v)],\\
 U(n,v)
 &=c_N\,p_{\theta,\mathcal C}(v\mid s,u,c_\Pi)
   \frac{\sqrt{N(n)}}{1+N(n,v)}.
\end{aligned}
 \label{eq:pucb}
\end{equation}
Here $n=(s,u)$ is a search node containing symbolic state $s$ and current token prefix $u$, while $v$ is a candidate next token. $Q(n,v)$ is the value estimate, $p_{\theta,\mathcal C}(v\mid s,u,c_\Pi)$ is the constrained model prior, $N(n)$ and $N(n,v)$ are parent and edge visit counts, and $c_N$ controls exploration. The second term is a policy-prior upper-confidence bound (PUCB) \citep{silver2017alphago}, a prior-weighted variant of UCT \citep{kocsis2006uct}. It lets the model order plausible continuations while search corrects locally likely choices with poor downstream effects.

Completed action sequences (or plans ($\pi$)) are evaluated using domain-independent planning heuristic:
\begin{equation}
  V(\pi)=
  \begin{cases}
    R_G, & s_\pi\models G,\\
    -h(s_\pi,G)-\lambda|\pi|, & \text{otherwise}.
  \end{cases}
  \label{eq:value}
\end{equation}
where, $s_\pi$ is the state reached by $\pi$, $R_G>0$ rewards the goal, $h(s_\pi,G)\geq0$ estimates remaining symbolic distance, and $\lambda>0$ penalizes plan length.  Standard planning heuristics instantiate $h$, so no learned critic is required.  Search stops at a goal or returns the highest-valued valid candidate within budget.  If the finite search tree contains a solution at token depth $d$, the constraints preserve every applicable action, and each permitted token has positive prior, PUCB visits the solution path with probability approaching one as the simulation budget grows; this is an asymptotic completeness statement, not a finite-budget guarantee.  PDDL supplies feasibility, the language model supplies semantic preference, and MCTS supplies global deliberation over their intersection.

\section{Experimental Setup}
\label{sec:experimental-setup}

\subsection{Benchmarks}

We evaluate in two complementary household simulators whose scenes span kitchens, living rooms, bedrooms, and bathrooms. \textbf{VirtualHome} represents activities as executable programs over objects in furnished Unity apartments \citep{puig2018virtualhome}. The dataset consists of 200 unique problems, testing novel object instances across four goal families: placing objects in a dishwasher, microwave, or refrigerator, and preparing food. These tasks primarily require multi-object rearrangement and correct reasoning about containment, support, proximity, and container state. \textbf{ALFWorld} aligns abstract household reasoning with visually embodied execution in AI2-THOR \citep{shridhar2021alfworld}. We evaluate all 134 episodes in its unseen split. The episodes cover six task families: pick-and-place, placing two objects, examining an object under a light, and cleaning, heating, or cooling an object before placement. Compared with VirtualHome, ALFWorld places greater emphasis on egocentric search, instance-level grounding, closed receptacles, and irreversible state transformations. Agents receive RGB observations and the task goal; simulator scene graphs and admissible-action lists are never exposed. For both benchmarks, a fixed rule-based parser converts the benchmark-provided symbolic goal annotation into $G$ using only goal classes and relations, not instance identities. It is applied uniformly, and the VLM has no access to the scene state, instance IDs, trajectories, or solutions.

\subsection{Models and Baselines}

Our main experiments use the Qwen3.5 family at 4B, 9B, and 27B parameters, allowing controlled comparison of model scale. We compare against direct interactive VLM policies, which predict one action from the current egocentric frame, and against Phase~II ablations that isolate reasoning, search, and formal constraints: unconstrained greedy decoding, unconstrained thinking, constrained greedy decoding, unconstrained MCTS, and constrained MCTS. All Phase~II comparisons reuse the same model-specific grounded initial state, so they differ only in planning. We additionally evaluate the open embodied models Embodied-R1.5 \citep{yuan2026embodiedr15}, MiMo-Embodied-7B \citep{xiaomi2025mimoembodied}, and RoboBrain2.0-32B \citep{robobrain2025technical}, together with API-based Gemini-3-Flash, as direct visual policies.

All methods receive the task goal, the same egocentric RGB interface, and the same simulator-generated outcome channel: success after a grounded action, or a concise error when the action is invalid or unreachable. Direct policies choose one action per turn. Our exploration harness additionally tracks inspection history and curates high-level exploration skills; this controller is part of the evaluated method and does not expose privileged scene state to the VLM. Phase~I decodes greedily and terminates when the task-relevant objects and relations are grounded or after $N_{\mathrm{exp}}$ steps. Interaction limits are 20 steps in VirtualHome and 55 in ALFWorld. Greedy conditions use temperature zero; thinking conditions use the model-recommended sampling parameters and a 32K-token context. 

MCTS evaluates nonterminal prefixes with domain-independent planning heuristics. We use a weighted average of common heuristics such as Fast Forward, LM-cut, and remaining plan cost by Downward~\citep{helmert2006fastdownward, hoffmann2001ff,helmert2009lmcut}. Additional details are presented in Appendix~\ref{app:extended-setup}.

\subsection{Metrics}

The primary metric is task success, which is one only when the goal state is reached as defined by the environment's simulator. We also report generated tokens, model-visible images, and failure attribution, and use successful plan length to characterize benchmark difficulty. Generated-token cost sums output tokens across all model calls for an episode. Image cost counts RGB frames provided to the model, rather than frames rendered internally by the simulator. Failure attribution assigns each unsuccessful episode to the first causal stage that prevents task completion.

\section{Results}
\label{sec:results}

We consider four core research questions a) whether factorizing visual perception and planning improves task completion over direct visual policies and unconstrained reasoning models; b) how inference-time search and token-level constraints interact to drive plan validity and goal reachability; c) whether these performance gains reduce generated-token and environment-observation costs; and finally d) how residual failures are causally distributed across visual state acquisition, action grounding, and search. We present the headline results here and note that we provide additional results in Appendix~\ref{app:additional-results}. 

\subsection{Overall Task Success}
\label{sec:results-main}

\begin{table}[htbp]
\centering
\small
\setlength{\tabcolsep}{1mm}
\begin{tabular}{llrr}
\toprule
Approach & Model & VH & ALFW \\
\midrule
\multicolumn{4}{l}{\emph{Neurosymbolic embodied agent}} \\
Ours & Qwen3.5-4B  & 94.5 & 90.3 \\
Ours & Qwen3.5-9B  & \textbf{99.5} & \textbf{97.8} \\
Ours & Qwen3.5-27B & \textbf{99.0} & 95.5 \\
\midrule
\multicolumn{4}{l}{\emph{Direct Qwen policies}} \\
Direct VLM & Qwen3.5-4B  & 5.5 & 1.5 \\
Direct VLM & Qwen3.5-9B  & 10.0 & 6.0 \\
Direct VLM & Qwen3.5-27B & 62.5 & 26.9 \\
\midrule
\multicolumn{4}{l}{\emph{API policy}} \\
Direct VLM & Gemini-3-Flash & 91.0 & 50.8 \\
\midrule
\multicolumn{4}{l}{\emph{Embodied VLM policies}} \\
Embodied VLM & MiMo-Embodied-7B & 18.5 & 0.7 \\
Embodied VLM & Embodied-R1.5-8B & 1.0 & 6.0 \\
Embodied VLM & RoboBrain2-32B & 27.0 & 9.7 \\
\midrule
\multicolumn{4}{l}{\emph{Planning from the Phase-I representation}} \\
Reasoning Enabled & Qwen3.5-4B  & 47.0 & 5.6 \\
Reasoning Enabled & Qwen3.5-9B  & 31.5 & 10.8 \\
Reasoning Enabled & Qwen3.5-27B & 59.0 & 16.9 \\
\bottomrule
\end{tabular}
\caption{Task success (\%) on VirtualHome and ALFWorld. Direct VLM and embodied VLM policies predict one environment action per interaction step. Best results are bold.}
\label{tab:main-results}
\end{table}

Table~\ref{tab:main-results} reports end-to-end task success. The proposed method reaches 94.5--99.5\% in VirtualHome and 90.3--97.8\% in ALFWorld, substantially outperforming direct visual interaction and unconstrained chain-of-thought reasoning\footnote{frequently referred to as thinking mode} across all model scales. These results show that perception and long-horizon planning, which typically compound errors when entangled in a single autoregressive policy, now become tractable when factorized into task-directed state acquisition and executable symbolic search.

The high success rates are consistent with this division as Phase~I asks the VLM to recover only a small set of task-relevant facts from targeted observations, producing a grounded PDDL problem file ($s_0, \beta_\tau$). Phase~II then solves an explicit symbolic problem rather than maintaining visual state and action semantics throughout a long interaction. The resulting agent is neurosymbolic in a functional sense. The VLM contributes visual recognition and commonsense priors, the transition model defines admissible state change, and MCTS deliberates over the intersection. None of these components alone must learn the complete embodied policy.

The factorization is also markedly parameter efficient. Our 4B agent obtains 94.5\% and 90.3\% success rates, exceeding the 27B direct VLM by 32.0\% and 63.4\% points in VirtualHome and ALFWorld, respectively. It likewise exceeds unconstrained thinking with the 27B model by 35.5 and 73.4 points. Thus, we see that increasing model capacity or allowing additional free-form reasoning does not substitute for an executable action space. A small model operating over a grounded transition system is more reliable than a substantially larger model required to infer perception, state tracking, action semantics, and long-horizon control jointly.

Gemini-3-Flash is the strongest direct-policy baseline, achieving 91.0\% in VirtualHome and 50.8\% in ALFWorld. We also note that the embodied-specialized baselines show limited transfer in this zero-shot setting. MiMo-Embodied-7B, Embodied-R1.5-8B, and RoboBrain2-32B achieve low success rates. In our setup, we are evaluating the zero-shot control through previously unseen simulator action interfaces under strict end-state goal verification. For example, MiMo-Embodied is trained across planning, affordance, and spatial tasks, Embodied-R1.5 introduces its own Planner--Grounder--Corrector framework, and RoboBrain2 targets spatial understanding and temporal decision-making~\citep{xiaomi2025mimoembodied,yuan2026embodiedr15,robobrain2025technical}. This result highlights the gap between pre-trained embodied priors and maintaining an executable, stateful policy over dozens of interaction turns~\citep[see][for more details]{he2025vlmformalize}. We furnish additional results in Appendix~\ref{app:additional-results}. 

\subsection{Inference and Environment Cost}
\label{sec:results-efficiency}

\begin{figure}[htbp]
\centering
\includegraphics[width=\columnwidth]{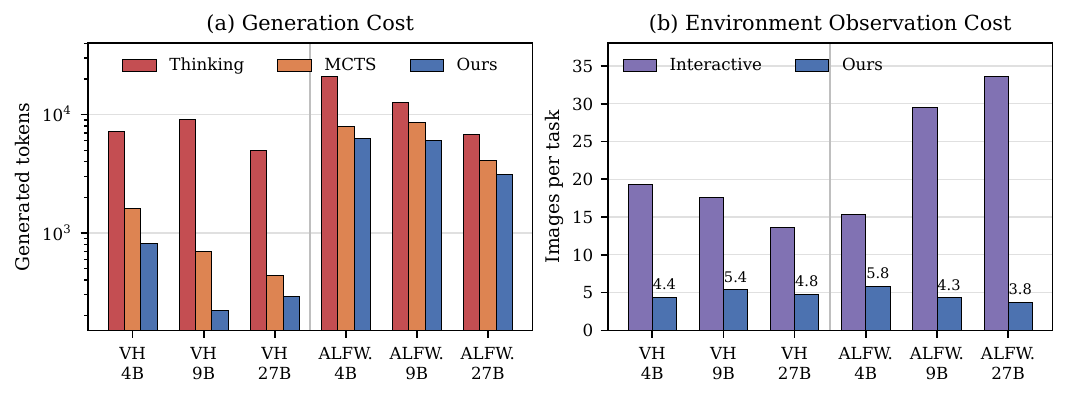}
\caption{Inference and environment-observation cost across model scales. (a) Mean generated tokens per task for unconstrained thinking, unconstrained MCTS, and our constrained MCTS. (b) Mean model-visible images per task for direct interactive policies and our two-phase agent. Lower is better in both panels.}
\label{fig:efficiency}
\end{figure}

Figure~\ref{fig:efficiency} shows that reliability does not necessitate higher inference or environment interaction costs.
We compare methods at matched model scale and average their costs across the two benchmarks. Under this paired comparison, constrained MCTS uses up to 1.5$\times$ fewer generated tokens than unconstrained MCTS and 4.0$\times$ fewer than extended thinking. Constraints remove inapplicable branches before the model spends tokens expanding them; search is then concentrated on executable alternatives rather than used as repeated generate-and-reject sampling.

The reduction in visual environment queries is even more pronounced.
Using the same fixed-scale averaging setup, our agent requires up to 5.5$\times$ fewer model-visible images than direct interaction, typically four to six rather than 13--34. Phase I obtains a compact task-relevant representation once, after which Phase II compares complete candidate plans without repeatedly querying the visual environment. This amortizes perception across the search tree. Visual inputs incur encoder computation that is not reflected in generated-token counts, so token-only comparisons understate the saving. In physical deployment, fewer perception--decision cycles also imply fewer environment queries and less unnecessary exploratory motion, although image count is not identical to low-level motor-action count.

\subsection{Failure Analysis}
\label{sec:results-errors}

\begin{figure}[htbp]
\centering
\includegraphics[width=0.62\columnwidth]{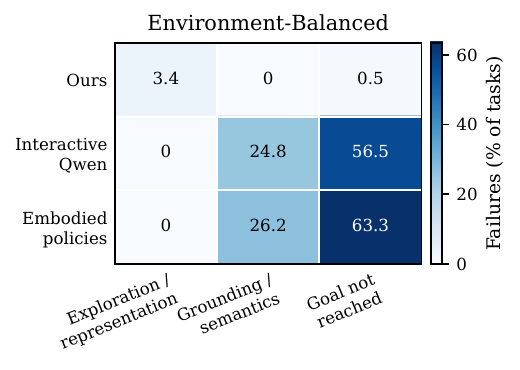}
\caption{
Terminal failure attribution, macro-averaged equally across VirtualHome and ALFWorld. Values are percentages of all tasks.
}
\label{fig:failure-attribution}
\end{figure}

We assign every unsuccessful episode to the earliest causal boundary that prevents completion. \emph{Exploration/representation} denotes an insufficient or incorrect Phase~I state, such as a missed goal object or an erroneous object--receptacle binding. \emph{Grounding/semantics} applies to direct policies whose proposed action cannot be executed with the intended effect because a precondition fails, an instance is incorrect or unreachable, or the symbolic action cannot be mapped to the simulator. \emph{Goal not reached} applies when execution remains possible but the goal is false at termination, including search- or interaction-budget exhaustion, loops, and premature \textsc{Done}.

We present our results in Figure~\ref{fig:failure-attribution}. We observe that the methods fail at different boundaries. Our residual error is concentrated upstream where we see that about 2.3\% of VirtualHome tasks and 4.5\% of ALFWorld tasks fail during representation acquisition, while only 1.0\% of ALFWorld tasks acquire a usable state but do not reach the goal. Once the symbolic state is accurate, constrained planning is therefore highly reliable. Qualitative inspection shows a characteristic visual failure where small, low-contrast objects, such as a fork blending into a tabletop, may remain unrecognized unless viewed at close range. The failure then propagates because the grounded initial state omits a required object. This is a limitation of the visual state-acquisition component rather than the applicability guarantee. We expect improvements in visual recognition and active viewpoint selection should transfer directly to the complete method.

Direct Qwen policies instead fail to reach the goal on 64.2\% of evaluated VirtualHome tasks. In ALFWorld, 39.8\% fail at grounding and a further 48.8\% terminate without satisfying the goal. The shift toward grounding in ALFWorld is consistent with its denser instance and receptacle semantics, where selecting a plausible object class is insufficient without binding the correct instance and interaction.

Embodied policies show the same qualitative bottleneck where failures concentrate in goal completion in VirtualHome and are divided between grounding and goal completion in ALFWorld. Table~\ref{tab:main-results} confirms that our proposal embeds action validity as an invariant within the constrained decoder and transition model, relieving the language model from inferring preconditions at each turn. As a result, failure modes isolate to state acquisition, with remaining errors stemming from incorrect visual representations rather than invalid plan generation. Additional results are in Appendices~\ref{app:additional-results} and~\ref{app:fine-grained-errors}.

\subsection{Constrained Planning Ablations}
\label{sec:results-ablation}

\begin{table}[htbp]
\centering
\small
\setlength{\tabcolsep}{1mm}
\begin{tabular}{lcccc}
\toprule
Method & Search & Constraints & VH & ALFW \\
\midrule
Greedy   & -- & -- & 61.0 & 15.4 \\
Thinking & -- & -- & 59.0 & 16.9 \\
Greedy   & -- & \checkmark & 98.5 & 32.3 \\
MCTS     & \checkmark & -- & \textbf{99.0} & 29.2 \\
Ours     & \checkmark & \checkmark & \textbf{99.0} & \textbf{95.5} \\
\bottomrule
\end{tabular}
\caption{Component ablation with Qwen3.5-27B. Entries report end-to-end task success (\%).}
\label{tab:component-ablation}
\end{table}

Table~\ref{tab:component-ablation} shows that neither additional generation nor either component in isolation explains the final result. In VirtualHome, constrained greedy decoding raises success from 61.0\% to 98.5\%, showing that applicability constraints account for most of the gain; MCTS without constraints reaches 99.0\%, but at the higher generation cost shown in Figure~\ref{fig:efficiency}. ALFWorld is more discriminating: constraints alone reach 32.3\% and unconstrained MCTS 29.2\%, whereas their combination reaches 95.5\%. The larger gap is consistent with a harder planning problem: successful 27B plans average 8.0 high-level actions in ALFWorld and 6.4 in VirtualHome, while the respective maxima are 30 and eight; ALFWorld also introduces more instance and receptacle bindings at each stage. Search and constraints are therefore complementary rather than interchangeable. Constraints make every explored prefix executable, while MCTS corrects locally plausible choices whose downstream consequences do not achieve the goal. Their nonlinear gain in ALFWorld supports the neurosymbolic design: symbolic semantics control feasibility, and model-guided search supplies goal-directed preference over the remaining valid plans.

\section{Conclusion}

We presented a neurosymbolic agent for long-horizon embodied task planning that separates task-directed visual exploration from constrained symbolic planning: a VLM and exploration harness construct a grounded initial state, while state-dependent decoding and MCTS generate a goal-directed plan under explicit PDDL dynamics. This factorization converts action applicability from a behavior the model must repeatedly infer into a constraint maintained throughout generation and search, enabling open 4B--27B models to substantially outperform larger direct visual policies, unconstrained reasoning, and embodied-specialized baselines while using fewer generated tokens and visual observations.

% bibliography is pre-compiled into main.bbl (arXiv does not run BibTeX)

\appendix
\section{Extended Experimental Setup}
\label{app:extended-setup}

\subsection{Evaluation Protocol}
We evaluate 200 VirtualHome tasks and all 134 episodes in the ALFWorld unseen split. Each agent receives the natural-language goal and egocentric RGB observations. The simulator scene graphs, object states, instance identifiers, and admissible actions are not exposed to the model. Environment adapters ground emitted class-level actions to simulator instances and return only whether the action executed successfully (e.g., \textsc{Previous action executed successfully}) or a concise failure reason when it is invalid or unreachable (e.g., \textsc{Your action could not be grounded from this pose}). All methods and baselines receive the same environment feedback. Note that, in our method and planning-from-Phase~I representation ablations, only Phase~I receives environment feedback while Phase~II does not. In comparison, direct-policy and visual-interaction baselines receive feedback at every environment step. Task success is determined exclusively from the environment simulation (e.g., Unity for VirtualHome).

Our method separates visual state acquisition from planning. Phase~I uses greedy decoding and a deterministic exploration harness to acquire goal-relevant predicates. It stops when the relevant state is grounded or after $N_{exp}$ steps. Phase~II receives the resulting symbolic handoff without any visual inputs and produces an open-loop plan of at most $N_{plan}$ actions, without access to the environment executor. $N_{exp} + N_{plan}$ is restricted to 20 in VirtualHome and 55 in ALFWorld.

This appendix documents the evaluation protocol and model hyperparameters used for all experiments.

\subsection{Direct Visual Interaction}
Direct policies emit one class-level action per turn from the current first-person observation. The environment parser accepts only actions in the benchmark-specific DSL, grounds the referenced class to an executable instance, advances the simulator after successful actions, and renders the next observation. Successful actions receive only a success acknowledgment, and failed actions receive a short grounding or reachability error and an unchanged observation. This is the same feedback Phase~I receives. The model must infer action postconditions and what action to execute next from the task description, feedback, updated image, and interaction history. Direct policies terminate on simulator success, an emitted \textsc{Done}, or the interaction limit: 20 steps in VirtualHome and 55 in ALFWorld. The interaction limit here is matched to $N_{exp} + N_{plan}$ in our method to ensure fairness. Direct Qwen, API, and embodied VLM policies all follow this same protocol.

\subsection{Sampling Parameters}
Table~\ref{tab:appendix-decoding} gives the decoding settings, following the usage recommended by the model providers. Phase~I, direct Qwen, greedy planning, constrained greedy planning, and both MCTS conditions use greedy decoding. The main-paper reasoning-enabled rows likewise report temperature-zero decoding. In this appendix, we further evaluate Qwen thinking using its recommended sampling settings over seeds 0, 42, and 1738. MiMo and Embodied-R1.5 use their recommended sampling configurations over three seeds (the main paper reports the seed-0 run). RoboBrain's main result uses its recommended greedy configuration, and we provide additional sensitivity analysis using nucleus sampling over three seeds here.

\begin{table*}[htbp]
\centering
\small
\setlength{\tabcolsep}{4pt}
\begin{tabular}{lccccc}
\toprule
Condition & Thinking & Temperature & Top-$p$ & Top-$k$ & Max output tokens \\
\midrule
Qwen Phase~I exploration & off & 0.0 & 1.0 & -- & 512  \\
Qwen direct interactive & off & 0.0 & 1.0 & -- & 128 per turn \\
Qwen greedy Phase~II & off & 0.0 & 1.0 & -- & 512 \\
Qwen MCTS Phase~II & off & 0.0 & 1.0 & -- & 128 per expansion \\
Qwen thinking, main paper & on & 0.0 & 1.0 & -- & 32,768 \\
Qwen thinking, supplementary & on & 1.0 & 0.95 & 20 & 32,768 \\
MiMo-Embodied-7B & off & 0.6 & 0.95 & -- & 512 per turn \\
Embodied-R1.5-8B & off & 0.7 & 0.8 & 20 & 512 per turn \\
RoboBrain2.0-32B, main paper & off & 0.0 & 1.0 & -- & 512 per turn \\
RoboBrain2.0-32B, supplementary & off & 1.0 & 0.9 & -- & 512 per turn \\
\bottomrule
\end{tabular}
\caption{Decoding settings. A dash denotes no top-$k$ cutoff. Qwen supplementary thinking, MiMo, Embodied-R1.5, and RoboBrain nucleus sampling are evaluated with seeds 0, 42, and 1738.}
\label{tab:appendix-decoding}
\end{table*}

\subsection{MCTS}
Both MCTS variants use a token-level prior-weighted form of UCT with expansion width 16, exploration constant 2.5, base constant 10, and early termination on a PDDL-valid goal plan. We restrict MCTS to 250 seconds per task. Hyperparameters were selected empirically. Constrained MCTS masks tokens that cannot extend an applicable PDDL action and caches reusable constrained prefixes. Unconstrained MCTS searches the full model distribution. Candidates are evaluated using a fixed weighted combination of domain-independent Fast Downward heuristics, including Fast Forward, LM-cut, and remaining plan cost. The same search settings are used for all models and benchmarks.

\subsection{Cluster Specifications}
Experiments used two GPU clusters with NVIDIA A100-80GB and NVIDIA H200 GPUs. Each evaluation job used one GPU, 8--16 CPU cores, and 128 GB RAM. Local-model inference used bfloat16 precision with vLLM.

\begin{table}[htbp]
\centering
\small
\setlength{\tabcolsep}{5pt}
\begin{tabular}{lrrr}
\toprule
Benchmark & Per episode & One run & Three runs \\
\midrule
VirtualHome & \$0.0148 & \$2.97 & \$8.90 \\
ALFWorld & \$0.0534 & \$7.16 & \$21.48 \\
\midrule
Combined & -- & \$10.13 & \$30.38 \\
\bottomrule
\end{tabular}
\caption{Estimated Gemini 3 Flash high-thinking API cost in USD.}
\label{tab:appendix-gemini-cost}
\end{table}

Table~\ref{tab:appendix-gemini-cost} shows the API costs for the Gemini evaluation runs. Local inference avoids per-token API charges, and our method is also more accurate.

\section{Statistical Analysis}
\label{app:statistics}

\subsection{Multiple Runs} We quantify stochastic sensitivity with three repeated runs; local-model runs use seeds 0, 42, and 1738. For each condition, we report the mean success rate, sample standard deviation across runs, and a 95\% percentile confidence interval from 10,000 task-cluster bootstrap replicates. Each replicate resamples task indices and applies the same indices to all three runs, preserving the dependence among repeated evaluations of the same task. Deterministic conditions are evaluated once.

\begin{table*}[htbp]
\centering
\small
\begin{tabular}{llrrrr}
\toprule
& & \multicolumn{2}{c}{VH} & \multicolumn{2}{c}{ALFW} \\
\cmidrule(lr){3-4}\cmidrule(lr){5-6}
Approach & Model & Accuracy (\%) & 95\% CI & Accuracy (\%) & 95\% CI \\
\midrule
Ours & Qwen3.5-4B & 94.5 $\pm$ 0.0 & [91.0, 97.5] & 90.3 $\pm$ 0.0 & [85.1, 94.8] \\
Ours & Qwen3.5-9B & \textbf{99.5 $\pm$ 0.0} & [98.5, 100.0] & \textbf{97.8 $\pm$ 0.0} & [94.8, 100.0] \\
Ours & Qwen3.5-27B & \textbf{99.0 $\pm$ 0.0} & [97.5, 100.0] & \textbf{95.5 $\pm$ 0.0} & [91.8, 98.5] \\
\midrule
Direct VLM & Qwen3.5-4B & 5.5 $\pm$ 0.0 & [2.5, 9.0] & 1.5 $\pm$ 0.0 & [0.0, 3.7] \\
Direct VLM & Qwen3.5-9B & 10.0 $\pm$ 0.0 & [6.0, 14.5] & 6.0 $\pm$ 0.0 & [2.2, 10.4] \\
Direct VLM & Qwen3.5-27B & 62.5 $\pm$ 0.0 & [55.5, 69.0] & 26.9 $\pm$ 0.0 & [19.4, 34.3] \\
\midrule
Qwen thinking & Qwen3.5-4B & 39.7 $\pm$ 3.1 & [34.2, 45.3] & 5.3 $\pm$ 0.5 & [2.4, 8.8] \\
Qwen thinking & Qwen3.5-9B & 31.0 $\pm$ 9.1 & [25.5, 36.8] & 14.1 $\pm$ 0.4 & [8.7, 20.0] \\
Qwen thinking & Qwen3.5-27B & 57.5 $\pm$ 7.9 & [51.5, 63.3] & 18.5 $\pm$ 0.8 & [12.3, 25.1] \\
\midrule
Embodied VLM & MiMo-Embodied-7B & 13.3 $\pm$ 4.5 & [10.5, 16.5] & 0.5 $\pm$ 0.4 & [0.0, 1.5] \\
Embodied VLM & Embodied-R1.5-8B & 0.3 $\pm$ 0.6 & [0.0, 0.8] & 4.0 $\pm$ 1.9 & [1.5, 7.0] \\
Embodied VLM & RoboBrain2.0-32B & 27.0 $\pm$ 0.0 & [21.0, 33.0] & 9.7 $\pm$ 0.0 & [5.2, 14.9] \\
\midrule
RoboBrain nucleus & RoboBrain2.0-32B & 0.0 $\pm$ 0.0 & [0.0, 0.0] & 3.5 $\pm$ 1.1 & [1.7, 5.5] \\
\midrule
API VLM (high thinking) & Gemini 3 Flash & 87.2 $\pm$ 0.8 & [83.7, 90.3] & 55.0 $\pm$ 1.6 & [47.3, 62.4] \\
\bottomrule
\end{tabular}
\caption{Accuracy and robustness across model conditions. The value after $\pm$ is the sample standard deviation across three runs. Deterministic conditions are marked $\pm 0.0$. Intervals are 95\% percentile task-cluster bootstrap intervals with repeated-run outcomes held together within each resample. Bold denotes the highest accuracy in each benchmark and results not significantly different from it.}
\label{tab:stochastic-robustness}
\end{table*}

Table~\ref{tab:stochastic-robustness} reports deterministic and stochastic conditions in a common format. The deterministic rows have zero decoding variance. Recommended-sampling Qwen thinking remains substantially below our method at every scale and on both benchmarks. Nucleus sampling does not improve RoboBrain's VirtualHome result and remains weak in ALFWorld. Among local conditions using neither constraints nor search, Qwen3.5-27B thinking is strongest, reaching 57.5\% in VirtualHome and 18.5\% in ALFWorld. The main paper reports a single Gemini 3 Flash run. Here, we additionally rerun Gemini 3 Flash three times with high thinking, reaching, on average, 87.2\% in VirtualHome and 55.0\% in ALFWorld.

Gemini uses 11.0 images per VirtualHome episode on average across the three runs, increasing to 23.2 images in ALFWorld. By comparison, the 27B method uses 4.8 and 3.8 model-visible images in Phase~I, after which Phase~II plans without further image-conditioned model calls. The factorized method therefore requires fewer image-conditioned interactions, and the final plan is executed in a single open-loop pass.

\subsection{Paired Significance Testing} We use exact paired McNemar tests for task-level success. When one condition is stochastic, it is paired against the deterministic condition separately for each run, and we retain the largest $p$-value, yielding a conservative test across runs. Holm correction controls the family-wise error rate at $\alpha=0.05$. We define two hypothesis families: (1) main comparisons, covering our method against direct policies, unconstrained thinking, embodied and API policies, and model-scale comparisons, and (2) component ablations, covering the 27B comparisons against constrained greedy and unconstrained MCTS. Holm correction is applied separately within each family.

\begin{table*}[htbp]
\centering
\small
\begin{tabular}{p{0.52\textwidth}rrrr}
\toprule
& \multicolumn{2}{c}{VH} & \multicolumn{2}{c}{ALFW} \\
\cmidrule(lr){2-3}\cmidrule(lr){4-5}
Comparison & Holm $p$ & Sig. & Holm $p$ & Sig. \\
\midrule
Ours (Qwen3.5-4B) vs. Direct VLM (Qwen3.5-4B) & $<0.001$ & \checkmark & $<0.001$ & \checkmark \\
Ours (Qwen3.5-9B) vs. Direct VLM (Qwen3.5-9B) & $<0.001$ & \checkmark & $<0.001$ & \checkmark \\
Ours (Qwen3.5-27B) vs. Direct VLM (Qwen3.5-27B) & $<0.001$ & \checkmark & $<0.001$ & \checkmark \\
Ours (Qwen3.5-4B) vs. Qwen thinking (Qwen3.5-4B) & $<0.001$ & \checkmark & $<0.001$ & \checkmark \\
Ours (Qwen3.5-9B) vs. Qwen thinking (Qwen3.5-9B) & $<0.001$ & \checkmark & $<0.001$ & \checkmark \\
Ours (Qwen3.5-27B) vs. Qwen thinking (Qwen3.5-27B) & $<0.001$ & \checkmark & $<0.001$ & \checkmark \\
Ours (Qwen3.5-4B) vs. Direct VLM (Qwen3.5-27B) & $<0.001$ & \checkmark & $<0.001$ & \checkmark \\
Ours (Qwen3.5-4B) vs. Qwen thinking (Qwen3.5-27B) & $<0.001$ & \checkmark & $<0.001$ & \checkmark \\
Ours (Qwen3.5-9B) vs. Embodied VLM (RoboBrain2.0-32B) & $<0.001$ & \checkmark & $<0.001$ & \checkmark \\
Ours (Qwen3.5-4B) vs. API VLM (high thinking) (Gemini 3 Flash) & 0.123 & -- & $<0.001$ & \checkmark \\
Ours (Qwen3.5-4B) vs. Ours (Qwen3.5-9B)  & 0.025 & \checkmark & 0.010 & \checkmark \\
Ours (Qwen3.5-9B) vs. Ours (Qwen3.5-27B) & 1.000 & --         & 0.500 & --         \\
\bottomrule
\end{tabular}
\caption{Exact paired McNemar tests for the main comparisons ($m = 24$ tests), with Holm correction within the family. For stochastic conditions, correction is applied to the maximum exact $p$-value across runs.}
\label{tab:mcnemar-main-comparisons}
\end{table*}

\begin{table}[htbp]
\centering
\small
\setlength{\tabcolsep}{4pt}
\begin{tabular}{@{}lrcrc@{}}
\toprule
& \multicolumn{2}{c}{VirtualHome} & \multicolumn{2}{c}{ALFWorld} \\
\cmidrule(lr){2-3}\cmidrule(lr){4-5}
Comparison & Holm $p$ & Sig. & Holm $p$ & Sig. \\
\midrule
Ours vs. Constrained greedy & 1.000 & -- & $<0.001$ & \checkmark \\
Ours vs. Unconstrained MCTS & 1.000 & -- & $<0.001$ & \checkmark \\
\bottomrule
\end{tabular}
\caption{Exact paired McNemar tests for the component ablations ($m = 4$ tests). All conditions use Qwen3.5-27B. Holm correction is applied within the family.}
\label{tab:mcnemar-component-ablations}
\end{table}

As shown in Table~\ref{tab:mcnemar-main-comparisons}, after correction, our method significantly outperforms the matched direct Qwen and recommended-sampling thinking baselines at every model scale in both environments. The 4B method also significantly exceeds the 27B direct and thinking baselines. Against Gemini 3 Flash with high thinking, the 4B method is not significantly different after Holm correction in VirtualHome and is significantly stronger in ALFWorld. Scaling our method from 4B to 9B is significant in both environments after correction, while the difference between 9B and 27B is not significant in either. 

In the ablations (see Table~\ref{tab:mcnemar-component-ablations}), constrained greedy and unconstrained MCTS are not significantly different from our method after Holm correction in VirtualHome; both are significantly weaker in ALFWorld, where combining search with transition-model constraints provides the decisive gain. The paired tests substantiate the paper's central task-accuracy claims.

\section{Additional Results}
\label{app:additional-results}

\subsection{Failure Attribution by Model and Environment}

\begin{figure*}[htbp]
\centering
\includegraphics[width=0.96\textwidth]{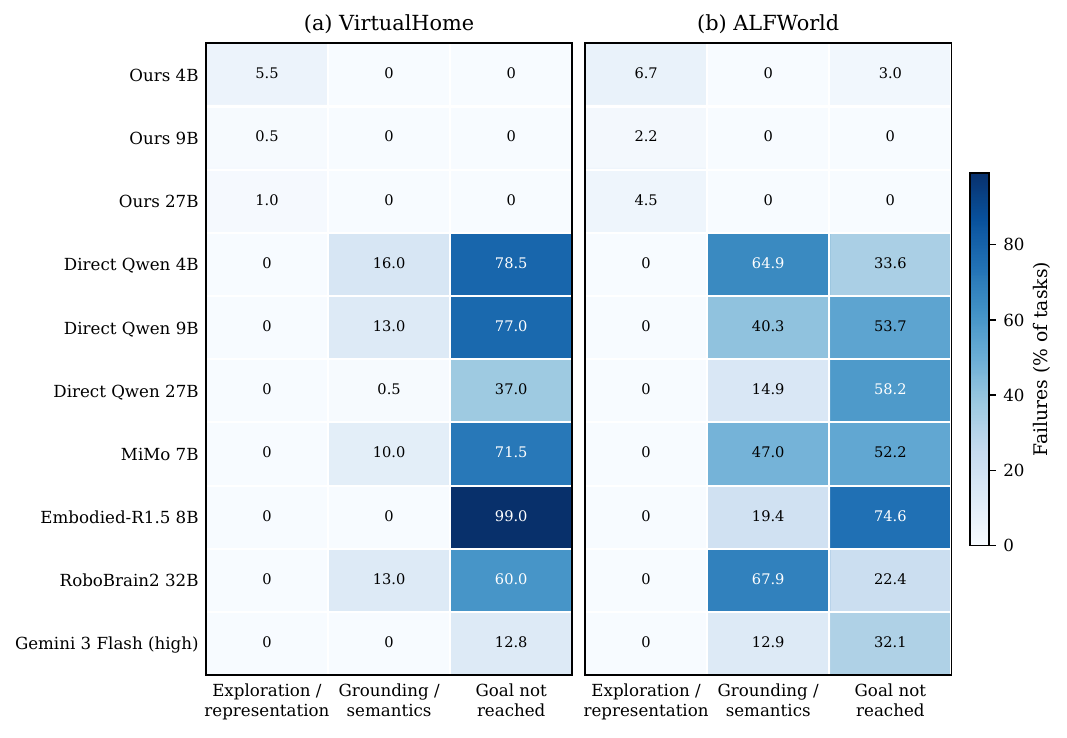}
\caption{Terminal failure attribution without aggregation across environments, Qwen scales, or visual-policy models. Values are percentages of all tasks for the corresponding row. Successful tasks contribute to none of the failure categories.}
\label{fig:appendix-failures}
\end{figure*}

Figure~\ref{fig:appendix-failures} resolves terminal failures by environment and by model, including the embodied policies and the Gemini API baseline, providing a finer-grained breakdown than the grouped results presented in the main text.

\paragraph{Discussion.}
The disaggregated results expose two trends hidden by the main-paper average. First, increasing the direct Qwen model scale reduces terminal grounding failures: from 16.0\% to 0.5\% in VirtualHome and from 64.9\% to 14.9\% in ALFWorld. However, the 27B policy still terminates without reaching the goal on 37.0\% and 58.2\% of tasks, respectively. Scale therefore improves action grounding more than long-horizon completion. Second, the embodied models fail at different boundaries despite similarly low overall accuracy. Embodied-R1.5 predominantly emits executable trajectories that do not complete the goal, whereas MiMo and RoboBrain encounter substantially more grounding failures in ALFWorld. Across Gemini's three high-thinking runs, 12.8\% of VirtualHome evaluations reach the interaction limit. In ALFWorld, 12.9\% terminate after repeated grounding failures, and 32.1\% terminate without reaching the goal. Our method's failures remain concentrated in representation acquisition from Phase~I, and only the ALFWorld 4B introduces a search-budget failure.

\subsection{Phase-II Results across Model Scales}

The main paper reports the 27B component ablation. Table~\ref{tab:all-p2-results} gives task success for every Qwen scale and Phase-II condition. The supplementary thinking row uses the recommended sampling configuration and reports the mean across three seeds, and the temperature-zero row is the single-run condition reported in the main paper.

\begin{table*}[htbp]
\centering
\small
\setlength{\tabcolsep}{2.5pt}
\begin{tabular}{lrrrrrr}
\toprule
& \multicolumn{3}{c}{VirtualHome} & \multicolumn{3}{c}{ALFWorld} \\
\cmidrule(lr){2-4}\cmidrule(lr){5-7}
Condition & 4B & 9B & 27B & 4B & 9B & 27B \\
\midrule
Unconstrained greedy & 66.0 $\pm$ 0.0 & 59.0 $\pm$ 0.0 & 61.0 $\pm$ 0.0 & 2.4 $\pm$ 0.0 & 10.0 $\pm$ 0.0 & 15.4 $\pm$ 0.0 \\
Thinking, temperature zero & 47.0 $\pm$ 0.0 & 31.5 $\pm$ 0.0 & 59.0 $\pm$ 0.0 & 5.6 $\pm$ 0.0 & 10.8 $\pm$ 0.0 & 16.9 $\pm$ 0.0 \\
Thinking, recommended sampling & 39.7 $\pm$ 3.1 & 31.0 $\pm$ 9.1 & 57.5 $\pm$ 7.9 & 5.3 $\pm$ 0.5 & 14.1 $\pm$ 0.4 & 18.5 $\pm$ 0.8 \\
Constrained greedy & 84.0 $\pm$ 0.0 & \textbf{99.5 $\pm$ 0.0} & \textbf{98.5 $\pm$ 0.0} & 24.0 $\pm$ 0.0 & 31.5 $\pm$ 0.0 & 32.3 $\pm$ 0.0 \\
Unconstrained MCTS & 78.0 $\pm$ 0.0 & 93.0 $\pm$ 0.0 & \textbf{99.0 $\pm$ 0.0} & 4.8 $\pm$ 0.0 & 22.3 $\pm$ 0.0 & 29.2 $\pm$ 0.0 \\
Ours & \textbf{94.5 $\pm$ 0.0} & \textbf{99.5 $\pm$ 0.0} & \textbf{99.0 $\pm$ 0.0} & \textbf{90.3 $\pm$ 0.0} & \textbf{97.8 $\pm$ 0.0} & \textbf{95.5 $\pm$ 0.0} \\
\bottomrule
\end{tabular}
\caption{Task success (\%) for every Qwen3.5 model scale and Phase-II condition, reported as accuracy $\pm$ standard deviation. Recommended-sampling thinking values use three seeds, and all other conditions use deterministic decoding. Bold denotes the best result in each column and any result not significantly different from it.}
\label{tab:all-p2-results}
\end{table*}

\paragraph{Discussion.}
The full scale breakdown reinforces the main ablation. Constraints alone are highly effective in VirtualHome, particularly for 9B and 27B, but neither constraints nor unconstrained search closes the ALFWorld gap. Their combination raises ALFWorld success to 90.3--97.8\% across all three scales. Recommended sampling changes the unconstrained-thinking result but does not alter this ordering.

\section{Fine-Grained Failure Analysis}
\label{app:fine-grained-errors}

The preceding attribution separates failures by causal stage. We further inspect the terminal mechanism recorded by the interaction harness and audit representative traces. Table~\ref{tab:fine-grained-failures} reports mutually exclusive terminal outcomes for the direct and embodied policies. These labels describe how an episode ended, rather than automatically inferring visual causality.

\begin{table*}[htbp]
\centering
\small
\setlength{\tabcolsep}{4pt}
\begin{tabular}{lllrrr}
\toprule
Environment & Policy & Model & Grounding & Interaction limit & Model \textsc{Done} \\
\midrule
VirtualHome & Direct Qwen & 4B & 16.0 & 71.5 & 7.0 \\
VirtualHome & Direct Qwen & 9B & 13.0 & 77.0 & 0.0 \\
VirtualHome & Direct Qwen & 27B & 0.5 & 37.0 & 0.0 \\
VirtualHome & Embodied & MiMo-7B & 10.0 & 40.0 & 31.5 \\
VirtualHome & Embodied & R1.5-8B & 0.0 & 50.0 & 49.0 \\
VirtualHome & Embodied & RoboBrain-32B & 13.0 & 60.0 & 0.0 \\
\midrule
ALFWorld & Direct Qwen & 4B & 64.9 & 32.1 & 1.5 \\
ALFWorld & Direct Qwen & 9B & 40.3 & 35.1 & 18.7 \\
ALFWorld & Direct Qwen & 27B & 14.9 & 44.8 & 13.4 \\
ALFWorld & Embodied & MiMo-7B & 47.0 & 35.8 & 16.4 \\
ALFWorld & Embodied & R1.5-8B & 19.4 & 68.7 & 6.0 \\
ALFWorld & Embodied & RoboBrain-32B & 67.9 & 17.2 & 5.2 \\
\bottomrule
\end{tabular}
\caption{Fine-grained terminal failure mechanisms as percentages of all tasks for each designated main-paper run. Grounding counts terminal mechanisms only. A policy can encounter many failed actions (Table~\ref{tab:failure-recovery}) without terminating on the ungroundable threshold.}
\label{tab:fine-grained-failures}
\end{table*}

The finer categories clarify why increased scale does not eliminate direct-policy failures. For direct Qwen in VirtualHome, terminal grounding failures fall from 16.0\% at 4B to 0.5\% at 27B, while interaction-limit failures remain at 37.0\%. In ALFWorld, grounding failures similarly fall from 64.9\% to 14.9\%, but the 27B model still loses 44.8\% of all tasks to the interaction limit and another 13.4\% to premature model termination. Larger models therefore produce more executable actions without consistently converting those actions into completed long-horizon goals.

The embodied policies exhibit distinct recovery behaviors. Embodied-R1.5 reaches the interaction limit or emits \textsc{Done} on 99.0\% of VirtualHome tasks, whereas RoboBrain's dominant ALFWorld outcome is repeated ungroundable action generation (67.9\%). MiMo lies between these two, with both grounding and budget exhaustion contributing substantially in ALFWorld. The aggregate embodied-policy row in the main paper consequently combines qualitatively different limitations.

\subsection{Response to Interaction Failure}

Terminal categories do not reveal whether a policy changes strategy after an unsuccessful action. We therefore measure behavior following the first ungroundable or failed execution. Table~\ref{tab:failure-recovery} reports the frequency of such events, eventual task recovery, immediate repetition of the failed class-level request, and persistent action loops.

\begin{table*}[htbp]
\centering
\small
\setlength{\tabcolsep}{5pt}
\begin{tabular}{lllrrrr}
\toprule
Environment & Policy & Model & Encounters failure & Recovers task & Repeats request & Persistent loop \\
\midrule
VirtualHome & Direct Qwen & 4B & 64.5 & 5.4 & 74.5 & 45.5 \\
VirtualHome & Direct Qwen & 9B & 53.5 & 9.3 & 33.0 & 79.5 \\
VirtualHome & Direct Qwen & 27B & 29.0 & 37.9 & 8.3 & 9.5 \\
VirtualHome & Embodied & MiMo-7B & 95.0 & 17.4 & 59.6 & 43.5 \\
VirtualHome & Embodied & R1.5-8B & 72.5 & 0.7 & 4.6 & 58.0 \\
VirtualHome & Embodied & RoboBrain-32B & 93.0 & 22.0 & 62.1 & 50.0 \\
\midrule
ALFWorld & Direct Qwen & 4B & 73.9 & 1.0 & 71.8 & 94.8 \\
ALFWorld & Direct Qwen & 9B & 79.9 & 5.6 & 41.7 & 72.4 \\
ALFWorld & Direct Qwen & 27B & 75.4 & 25.7 & 18.4 & 42.5 \\
ALFWorld & Embodied & MiMo-7B & 96.3 & 0.0 & 63.3 & 87.3 \\
ALFWorld & Embodied & R1.5-8B & 70.9 & 5.3 & 52.1 & 92.5 \\
ALFWorld & Embodied & RoboBrain-32B & 93.3 & 7.2 & 84.3 & 85.1 \\
\bottomrule
\end{tabular}
\caption{Failure response of direct and embodied policies (\%). Encounters failure is the fraction of tasks containing an ungroundable or failed execution. Recovers task is success among those affected tasks. Repeats request is the fraction of failed-action transitions followed by the same class-level request. Persistent loop is the fraction of all tasks containing at least five identical consecutive requests or a six-step two-action oscillation.}
\label{tab:failure-recovery}
\end{table*}

Scaling direct Qwen substantially improves local recovery. In VirtualHome, the fraction of failed-action transitions followed by the same request falls from 74.5\% at 4B to 8.3\% at 27B, and among episodes that encounter an execution failure, eventual success rises from 5.4\% to 37.9\%. ALFWorld shows the same direction, although it remains harder: Qwen-27B recovers 25.7\% of affected episodes and still enters a persistent loop on 42.5\% of all tasks. The non-monotonic VirtualHome loop rate at 9B reflects a distinction between termination and persistence: weaker policies may exhaust the grounding-failure threshold early, whereas an executable but unproductive policy can continue until the larger step budget is exhausted.

The embodied models fail differently rather than uniformly. In VirtualHome, R1.5 rarely repeats the exact failed request (4.6\%) but frequently enters alternating-view or repeated-action loops (58.0\%), matching the rotation sequence in Figure~\ref{fig:reproduced-failure-sequences}. In ALFWorld, RoboBrain repeats the same class-level request after 84.3\% of failed transitions, while MiMo encounters at least one failed execution in 96.3\% of tasks and never recovers one of those affected episodes in the seed-0 run. Explicit negative feedback is therefore available but is not consistently converted into a revised grounded strategy.

\subsection{Representation-Level Errors}

An audit of VirtualHome method failures across model scales identifies several failures involving small objects (e.g., \texttt{pie}, \texttt{juice}), which are compact food objects whose class and supporting relation must be recovered from a wide household view. The model may identify a room and supporting surface without resolving the small goal object, or identify both entities without committing their support relation. In ALFWorld, long searches can instead visit the appropriate region of the scene and inspect many plausible receptacles without exposing the target before the exploration budget expires. Thus, the residual Phase-I errors separate into \emph{local perceptual ambiguity} and \emph{incomplete search coverage}, neither of which is captured by a single missing-object count.

\subsection{Representative Policy Traces}

\begin{figure*}[htbp]
\centering
\setlength{\tabcolsep}{2pt}
\begin{tabular}{c@{\hspace{2pt}}c@{\hspace{2pt}}c@{\hspace{2pt}}c}
\scriptsize 1. Agent spawns & \scriptsize 2. Inspects table & \scriptsize 3. Moves away from table & \scriptsize \textcolor{red!65!black}{4. Reaches counter, fork visually missed} \\
\includegraphics[width=.23\textwidth]{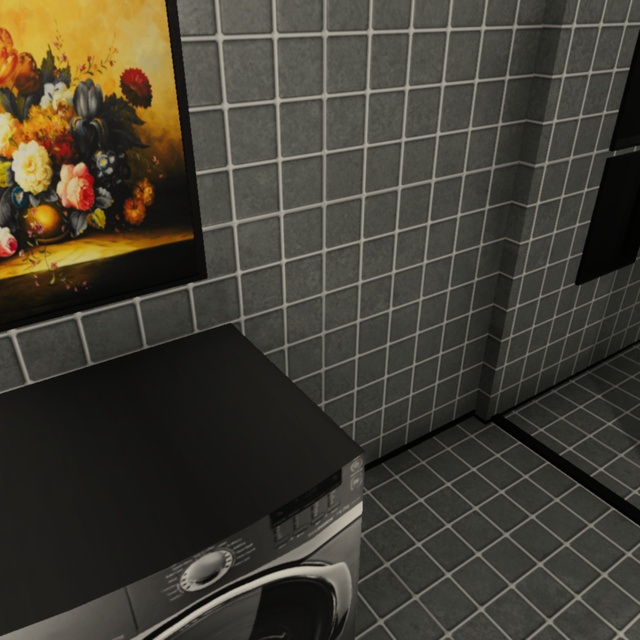} &
\includegraphics[width=.23\textwidth]{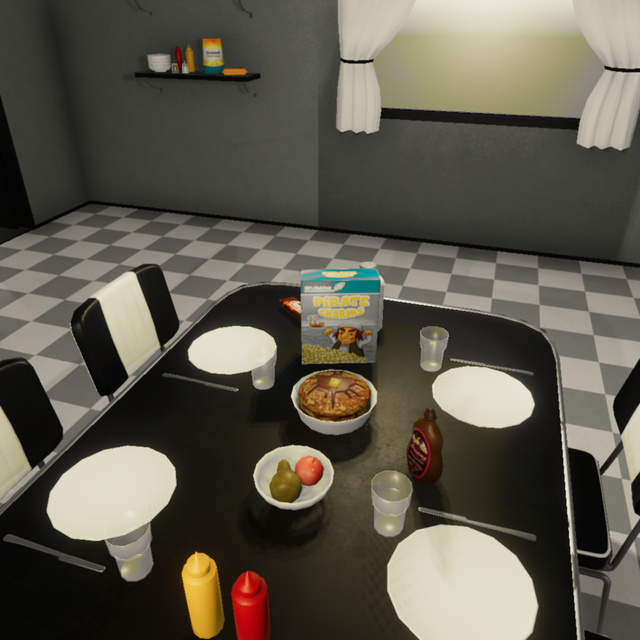} &
\includegraphics[width=.23\textwidth]{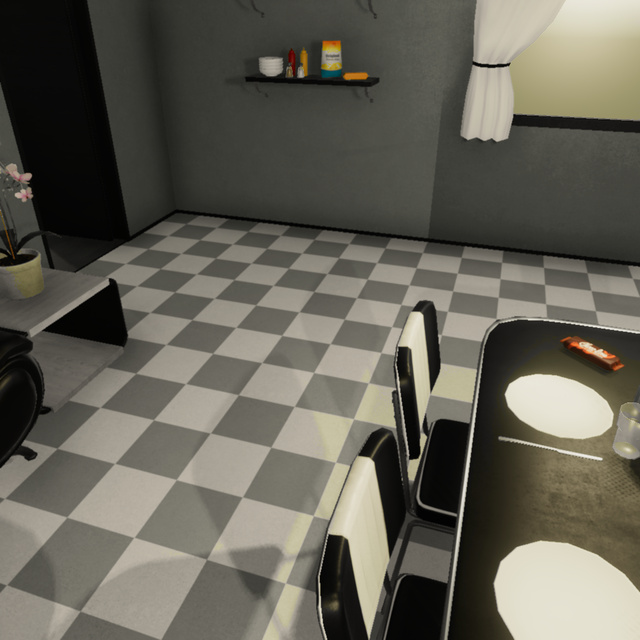} &
\includegraphics[width=.23\textwidth]{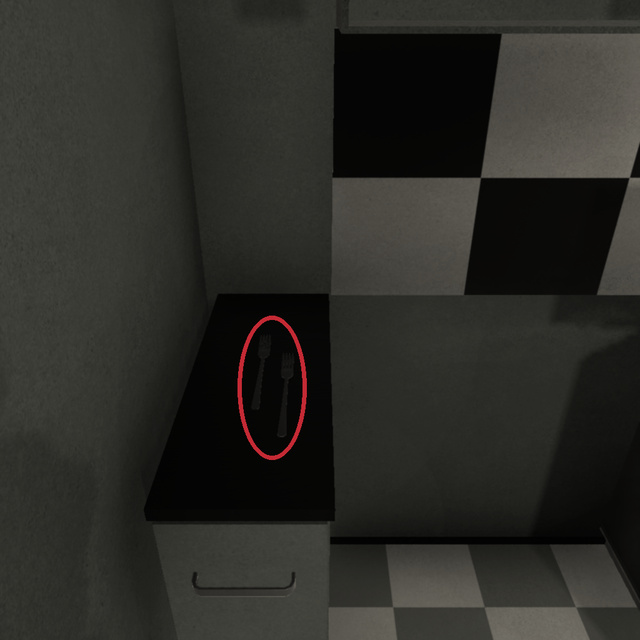} \\
\multicolumn{4}{l}{\small\textbf{(a) Method, VirtualHome:} directed exploration reaches the correct counter, but the visually subtle fork is not recovered.} \\[3pt]
\scriptsize 1. Agent spawns & \scriptsize 2. Finds first CD & \scriptsize 3. Searches other drawers & \scriptsize \textcolor{red!65!black}{4. Budget ends without second CD} \\
\includegraphics[width=.23\textwidth]{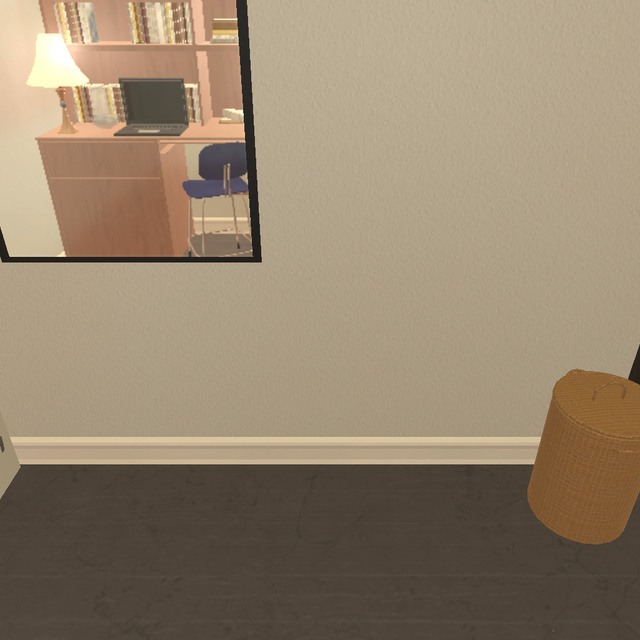} &
\includegraphics[width=.23\textwidth]{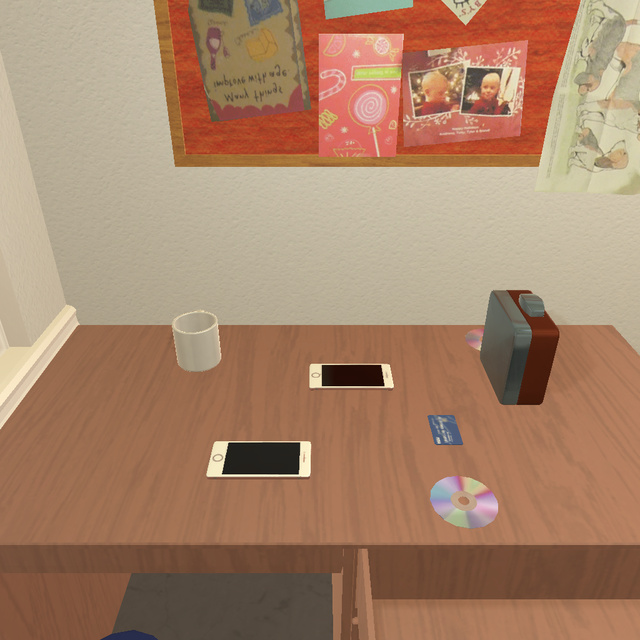} &
\includegraphics[width=.23\textwidth]{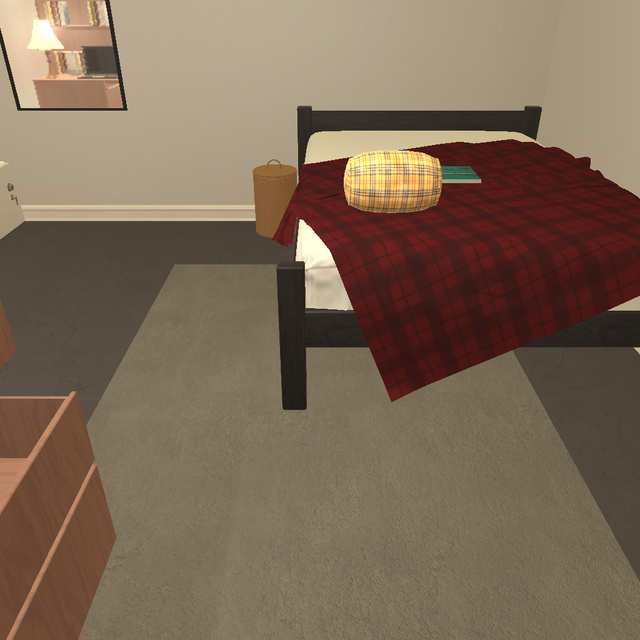} &
\includegraphics[width=.23\textwidth]{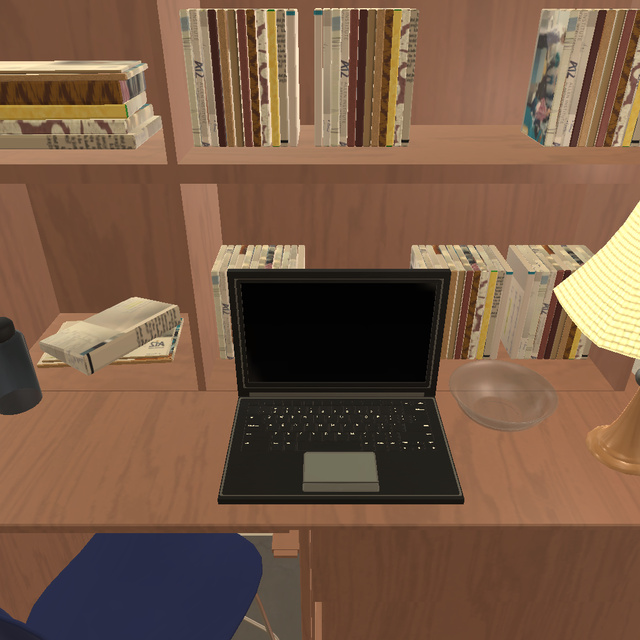} \\
\multicolumn{4}{l}{\small\textbf{(b) Method, ALFWorld:} one required CD is found, but active search exhausts its budget before localizing the second instance.} \\[3pt]
\scriptsize 1. Reaches refrigerator & \scriptsize 2. Opens it & \scriptsize \textcolor{red!65!black}{3. Closes it} & \scriptsize \textcolor{red!65!black}{4. Reopens it, loop continues} \\
\includegraphics[width=.23\textwidth]{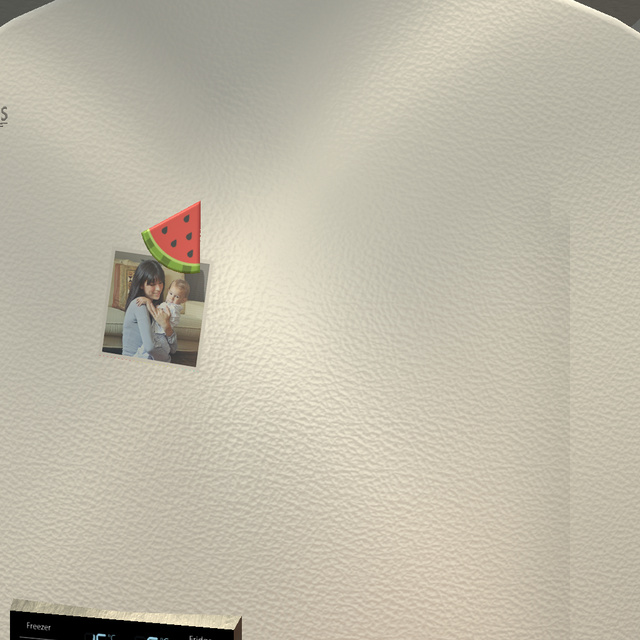} &
\includegraphics[width=.23\textwidth]{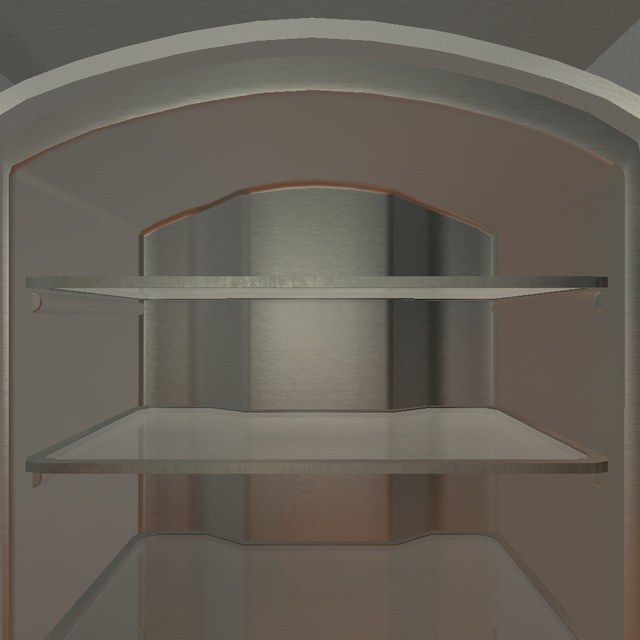} &
\includegraphics[width=.23\textwidth]{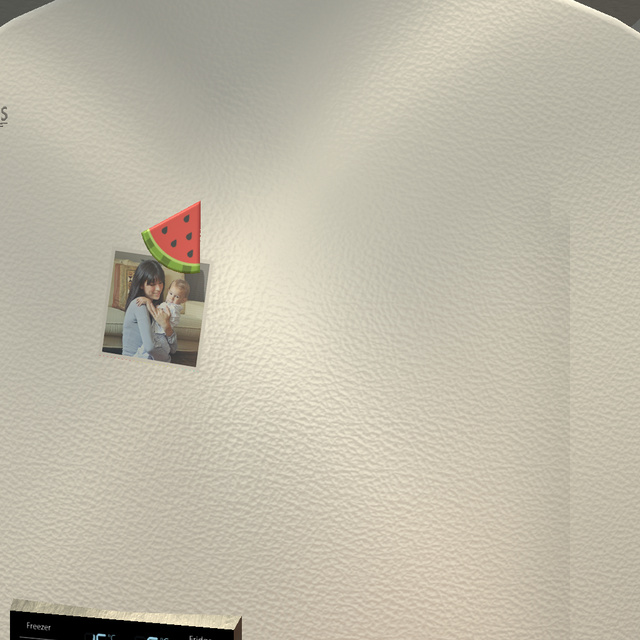} &
\includegraphics[width=.23\textwidth]{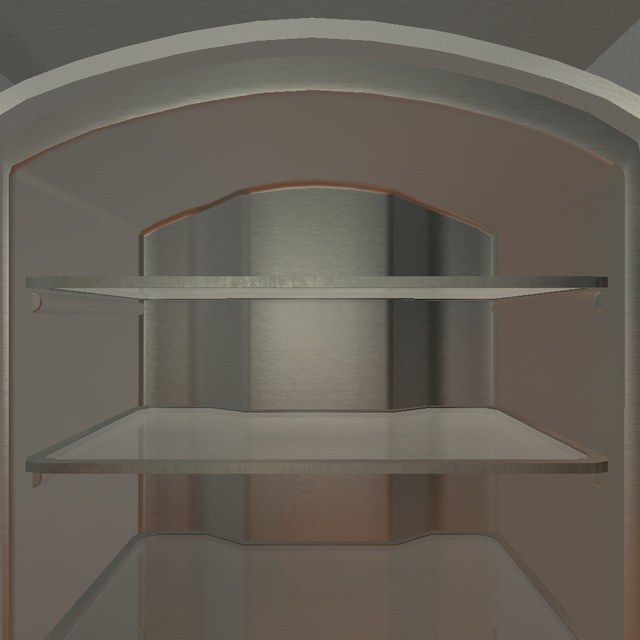} \\
\multicolumn{4}{l}{\small\textbf{(c) Direct Qwen, ALFWorld:} valid actions repeatedly reverse the same refrigerator state until the episode ends.} \\[3pt]
\scriptsize 1. Agent spawns & \scriptsize 2. Opens microwave & \scriptsize \textcolor{red!65!black}{3. Turns left after failed put} & \scriptsize \textcolor{red!65!black}{4. Turns back, loop begins} \\
\includegraphics[width=.23\textwidth]{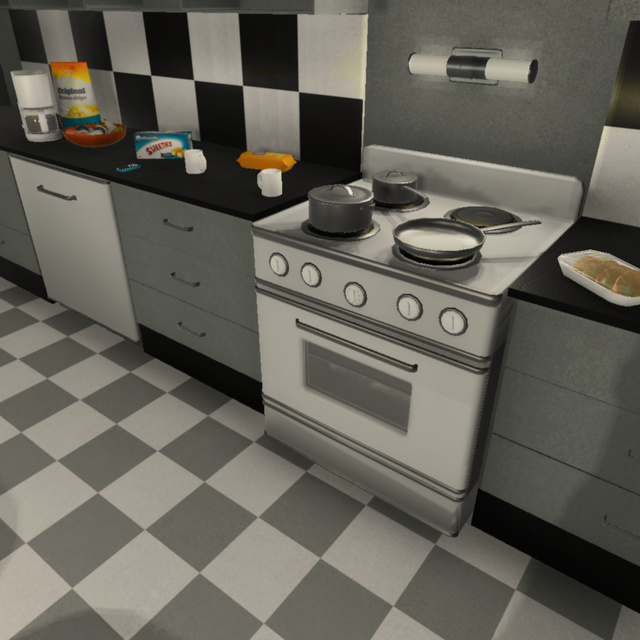} &
\includegraphics[width=.23\textwidth]{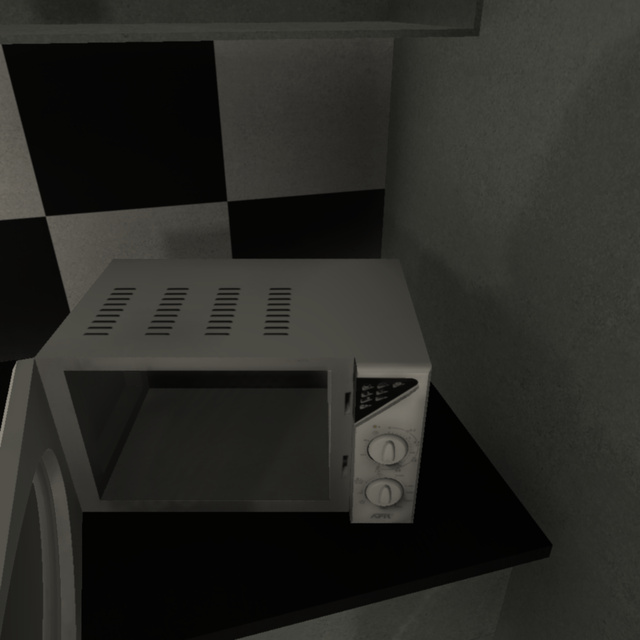} &
\includegraphics[width=.23\textwidth]{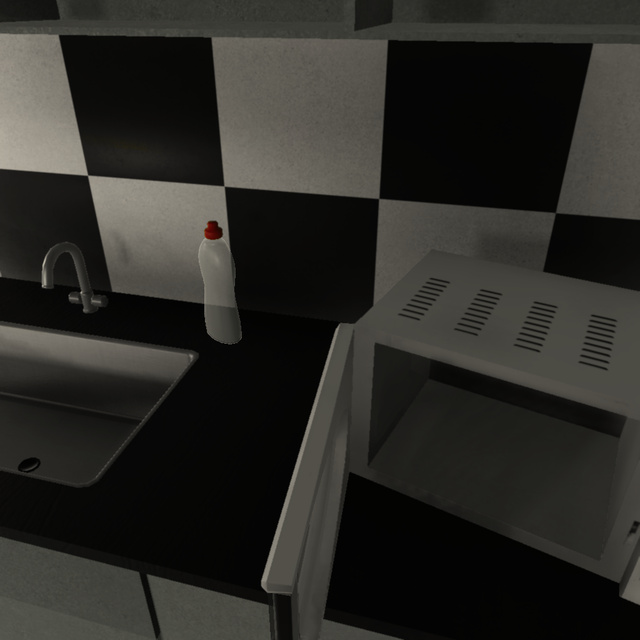} &
\includegraphics[width=.23\textwidth]{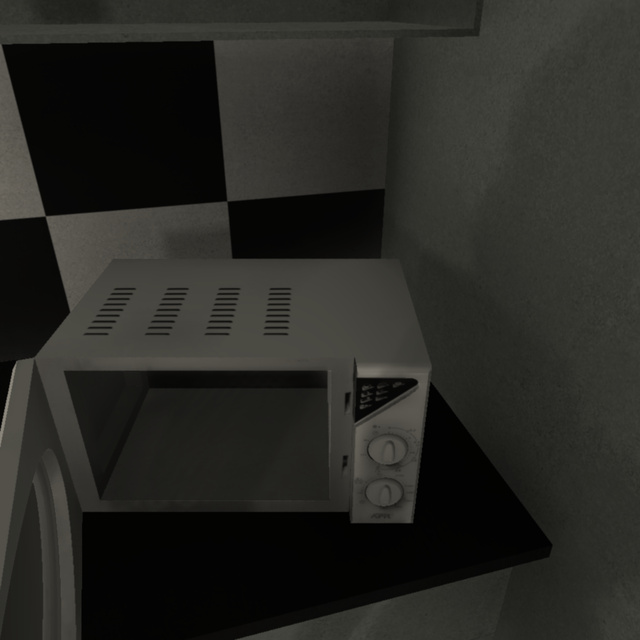} \\
\multicolumn{4}{l}{\small\textbf{(d) Embodied-R1.5, VirtualHome:} after reaching the appliance, manipulation failure devolves into view oscillation.}
\end{tabular}
\caption{Example failure sequences. Each row follows an episode from its initial or first informative state through progress to the decisive error and its terminal consequence. The red callout in (a) identifies the relevant object.}
\label{fig:reproduced-failure-sequences}
\end{figure*}

Figure~\ref{fig:reproduced-failure-sequences} illustrates four failure mechanisms with example episodes from each simulator and task family. All prevalence statements come from the complete recorded results in Table~\ref{tab:fine-grained-failures}.

Together, these examples separate four mechanisms that success rate alone conflates: local perception, finite active-search coverage, repeated state reversal, and view-control oscillation after an interaction failure. They also explain why scale alone is insufficient. A larger model may select a semantically appropriate receptacle or action more reliably, while still lacking the persistent scene memory, global progress state, or recovery policy required to complete the episode. The method moves much of that burden into an explicit handoff and planner, but remains bounded by what Phase I can expose within its exploration budget. Direct and embodied policies avoid a fixed symbolic bottleneck, yet must repeatedly solve grounding and progress tracking online.

\section{Prompts}
\label{app:prompts}

All prompts use the same templates across model scales. The examples below instantiate those templates for a VirtualHome task that places a wineglass in the dishwasher. ALFWorld uses similarly styled prompts. Figures~\ref{fig:appendix-phase1-prompt}--\ref{fig:appendix-interactive-prompt} show the instantiated prompts for each phase.

\subsection{Phase I: Task-Directed Exploration}

At each exploration step, the VLM receives a first-person view, the goal-relevant object classes, the accumulated belief, prior grounded actions, environment feedback, and a harness-curated set of currently available high-level inspection skills. A representative prompt is shown in Figure~\ref{fig:appendix-phase1-prompt}.

The complete template additionally defines the accepted predicate schema, benchmark class-name normalization, and the semantics of actions. The model may select only one of the high-level skills supplied for the current state. The harness grounds that skill, executes it, and accepts relational facts only when supported by the resulting observation or interaction.

\subsection{Phase II: Planning}

Phase~II receives the class-level action reference, the grounded state produced by Phase~I, and the symbolic goal. Instance identifiers remain hidden. No image is passed to Phase~II. A representative unconstrained prompt is shown in Figure~\ref{fig:appendix-phase2-prompt}.

The constrained conditions use the same natural-language prompt. Their only difference is at decoding time: after every completed action, the PDDL transition model updates the symbolic state and masks token continuations that cannot form an applicable next action.

\subsection{Direct Visual Interaction}

The direct Qwen policy begins with a prompt describing first-person observation, reachability, the action DSL, class-name conventions, and the requirement to emit exactly one action. Figure~\ref{fig:appendix-interactive-prompt} shows the first user turn, action vocabulary, environment feedback, and the history representation.

This feedback exposes neither a valid-action list nor a symbolic state. Successful feedback deliberately omits inferred postconditions, since the model must interpret the updated observation and reason from the action semantics. Embodied policies receive the same task and interaction setup, shown in Figure~\ref{fig:appendix-interactive-prompt}.

MiMo and RoboBrain return a bare action. Embodied-R1.5 returns the same action inside \texttt{<answer>... </answer>}. Thinking is disabled for all three embodied-specialized policies.

\begin{figure*}[htbp]
\centering
\begin{tikzpicture}
\node[fill=phOneBg, draw=phOne, rounded corners=3pt, inner sep=7pt,
      text width=0.92\textwidth, align=left, font=\footnotesize\ttfamily] {
You are a vision-language agent in VirtualHome, a household simulator.\\
You see one or more first-person RGB images. Your only job in this phase is to find the listed objects and determine where each one is.\\
Do not plan or execute the household chore. Reply with JSON only.\\[3pt]
\textbf{Task:}\\
Find each listed object from first-person vision and determine where it is.\\[3pt]
\textbf{Objects to find and locate:}\\
wineglass, dishwasher\\[3pt]
\textbf{Current belief:}\\
visible kitchen; visible kitchentable; visible dishwasher; closed dishwasher\\[3pt]
\textbf{Previous actions:}\\
1. goto kitchen -> success\\
2. goto kitchentable -> success\\[3pt]
\textbf{Valid skills right now:}\\
- goto wineglass\\
- look\\[3pt]
\textbf{Feedback:}\\
The previous action executed successfully.\\[3pt]
Return visible predicates, optional object locations, exactly one valid skill in ``action'', and ``done'': true only after every listed object has a grounded placement.
};
\end{tikzpicture}
\caption{Example VirtualHome Phase-I exploration prompt.}
\label{fig:appendix-phase1-prompt}
\end{figure*}

\begin{figure*}[htbp]
\centering
\begin{tikzpicture}
\node[fill=phTwoBg, draw=phTwo, rounded corners=3pt, inner sep=7pt,
      text width=0.92\textwidth, align=left, font=\footnotesize\ttfamily] {
You control an embodied agent in a simulated household called VirtualHome. Output a sequence of actions that changes the observable initial state into the goal state. An action takes effect only if all of its conditions already hold. The agent cannot act on an object it is not close to.\\[3pt]
\textbf{Actions:}\\
walk \textless object/room\textgreater: move close to an object or room\\
grab \textless object\textgreater: hold a nearby object\\
open \textless container\textgreater: open a nearby closed container\\
putin \textless object\textgreater{} \textless container\textgreater: place a held object in a nearby open container\\
DONE\\[3pt]
\textbf{Initial state:}\\
visible kitchen\\
visible kitchentable\\
visible dishwasher\\
on wineglass kitchentable\\
closed dishwasher\\
handempty\\[3pt]
\textbf{Goal:}\\
inside wineglass dishwasher\\[3pt]
Each action should make progress toward the goal and respect preconditions. Output one action per line using class names only, and stop immediately after the action that achieves the goal.\\[3pt]
\textbf{Plan:}
};
\end{tikzpicture}
\caption{Example VirtualHome Phase-II planning prompt.}
\label{fig:appendix-phase2-prompt}
\end{figure*}

\begin{figure*}[htbp]
\centering
\begin{tikzpicture}
\node[fill=handBg, draw=hand, rounded corners=3pt, inner sep=7pt,
      text width=0.92\textwidth, align=left, font=\footnotesize\ttfamily] {
\textbf{FIRST USER TURN}\\
Goal: Put the wineglass in the dishwasher.\\
Goal objects (exact VirtualHome class names): wineglass, dishwasher\\
Use the first-person image and choose the next action toward the goal.\\
Output only one action line (or DONE).\\[3pt]
\textbf{ACTION FORMS}\\
walk \textless object/room\textgreater\\
grab \textless object\textgreater\\
open \textless container\textgreater\\
close \textless container\textgreater\\
putback \textless object\textgreater{} \textless surface\textgreater\\
putin \textless object\textgreater{} \textless container\textgreater\\
turnleft; turnright; walkforward; DONE\\[3pt]
\textbf{SUCCESSFUL-ACTION FEEDBACK}\\
Previous action executed successfully.\\
Here is the updated first-person view. Choose the next action (or DONE).\\[3pt]
\textbf{FAILED-ACTION FEEDBACK}\\
Your action ``grab wineglass'' could not be grounded from this pose (invalid or unreachable).\\
The view did not change. Try a different next action.\\[3pt]
\textbf{ACTION HISTORY}\\
Previous actions:\\
1. walk wineglass -> success\\
2. grab wineglass -> success\\
3. walk dishwasher -> success\\
4. putin wineglass dishwasher -> failed (invalid or unreachable)
};
\end{tikzpicture}
\caption{VirtualHome direct-interaction prompt components and environment feedback. The current first-person image accompanies each request.}
\label{fig:appendix-interactive-prompt}
\end{figure*}

\end{document}